\documentclass[conference]{IEEEtran}
\IEEEoverridecommandlockouts

\usepackage{moreverb,cite}
\usepackage[colorlinks,bookmarksopen,bookmarksnumbered,citecolor=red,urlcolor=red]{hyperref}      
\usepackage{comment}
\usepackage[dvips]{graphicx}
\usepackage{amssymb, amsmath, amsfonts}  

\newtheorem{remark}{Remark}[section]

\usepackage{algorithm}
\usepackage{algpseudocode}

\renewcommand{\theequation}{\thesection.\arabic{equation}}

\def\BibTeX{{\rm B\kern-.05em{\sc i\kern-.025em b}\kern-.08em
		T\kern-.1667em\lower.7ex\hbox{E}\kern-.125emX}}
	
\def\be{\begin{equation}}
\def\bea{\begin{eqnarray}}
\def\beas{\begin{eqnarray*}}
\def\eea{\end{eqnarray}}
\def\eeas{\end{eqnarray*}}

\def\bi{\begin{itemize}}
\def\ee{\end{equation}}
\def\ei{\end{itemize}}

\def\ee{{\epsilon}}

\def\be{\begin{equation}}

\def\bi{\begin{itemize}}
\def\ee{\end{equation}}
\def\ei{\end{itemize}}

\def\z1{z^{-1}}

\newcommand{\Fcal}{\mathcal{F}}

\newcommand{\Ncal}{\mathcal{N}}
\newcommand{\Acal}{\mathcal{A}}

\newcommand{\Rsf}{\mathsf{R}}

\newcommand{\Rz}{\mathsf{R}_{\mathrm{0}}}
\newcommand{\Ro}{\mathsf{R}_{\mathrm{1}}}
\newcommand{\Ri}{\mathsf{R}_{\mathrm{i}}}
\newcommand{\prm}{\mathrm{p}}
\newcommand{\vrm}{\mathrm{v}}
\newcommand{\Msf}{\mathsf{M}}
\newcommand{\qrm}{\mathrm{q}}
\newcommand{\rrm}{\mathrm{r}}
\newcommand{\xrm}{\mathrm{x}}
\newcommand{\arm}{\mathrm{a}}
\newcommand{\urm}{\mathrm{u}}

\newcommand{\Qrm}{\mathrm{Q}}
\newcommand{\Grm}{\mathrm{G}}
\newcommand{\yrm}{\mathrm{y}}
\newcommand{\zrm}{\mathrm{z}}
\newcommand{\irm}{\mathrm{i}}
\newcommand{\Irm}{\mathrm{I}}
\newcommand{\xk}{\mathrm{x}_{k}}

\def\diag{\mbox{ diag } }

\def \benumroman{\renewcommand{\theenumi}{\roman{enumi}} \begin{enumerate}}
\def \eenumroman{\end{enumerate}}

\begin{document}
\title{Infrastructure-free Localization of Aerial Robots with Ultrawideband Sensors\\
	\thanks{The research reported in this publication was supported by funding from King Abdullah University of Science and Technology (KAUST).}
}

\author{\IEEEauthorblockN{Samet~G\"{u}ler,~Mohamed~Abdelkader,~and~Jeff~S.~Shamma}
	\thanks{$^{1}$Samet G\"{u}ler, Mohamed~Abdelkader, and Jeff S. Shamma are with Robotics, Intelligent Systems, and Control (RISC) Lab, Computer, Electrical and Mathematical Science and Engineering Division (CEMSE), KAUST, Thuwal 23955--6900, Saudi Arabia.
		{\tt\small samet.guler@kaust.edu.sa}}%
}

\maketitle

\begin{abstract}
Robots in a swarm take advantage of a motion capture system or GPS sensors to obtain their global position. However, motion capture systems are environment-dependent and GPS sensors are not reliable in occluded environments. For a reliable and versatile operation in a swarm, robots must sense each other and interact locally. Motivated by this requirement, here we propose an on-board localization framework for multi-robot systems. Our framework consists of an anchor robot with three ultrawideband (UWB) sensors and a tag robot with a single UWB sensor. The anchor robot utilizes the three UWB sensors as a localization infrastructure and estimates the tag robot's location by using its on-board sensing and computational capabilities solely, without explicit inter-robot communication. We utilize a dual Monte-Carlo localization approach to capture the agile maneuvers of the tag robot with an acceptable precision. We validate the effectiveness of our algorithm with simulations and indoor and outdoor experiments on a two-drone setup. The proposed dual MCL algorithm yields highly accurate estimates for various speed profiles of the tag robot and demonstrates a superior performance over the standard particle filter and the extended Kalman Filter.
\end{abstract}


\setcounter{equation}{0}
\section{Introduction}\label{sec:intro}
\IEEEPARstart{A}{utonomous} mobile robots have been deployed in various civil and military applications such as goods delivery in urban areas, service industry, manufacturing, and border security. A mobile robot should have a high-performance localization algorithm because a mobile robot's decision mechanism can function reliably only with a good positioning framework. The mobile robot localization problem is defined as developing a hypothesis about a robot's location in a given environment which is usually represented with a set of landmarks or a detailed map. In a multi-robot system, a mobile robot needs to localize itself with respect to the other robots as well.

The standard methods for mobile robot localization include geometric, optimization, and filtering methods. The geometric and optimization based approaches take a set of anchor-sensor distance measurements at a specific time instant and produce solutions for the possible sensor locations based on the distance geometry. The accuracy of both methods suffers from measurement noises and motion of the localized sensors. If the distance measurements are constantly available, then various filtering approaches can be used. The Bayesian approaches, particularly the extended Kalman filter (EKF), were commonly employed for localization. The filtering based methods first predict the robot motion with inertial sensors then update the belief with exteroceptive sensor data. EKF localization yields good performance in many scenarios. However, tuning the EKF parameters requires plenty of time and experiments, the initial condition significantly affects the EKF performance, and EKF localization usually does not suffice to track agile robot motions.

\begin{figure}[!b]
	\centering
	\includegraphics[trim = 0cm 0cm 0cm 0cm, clip, width=0.35\textwidth]{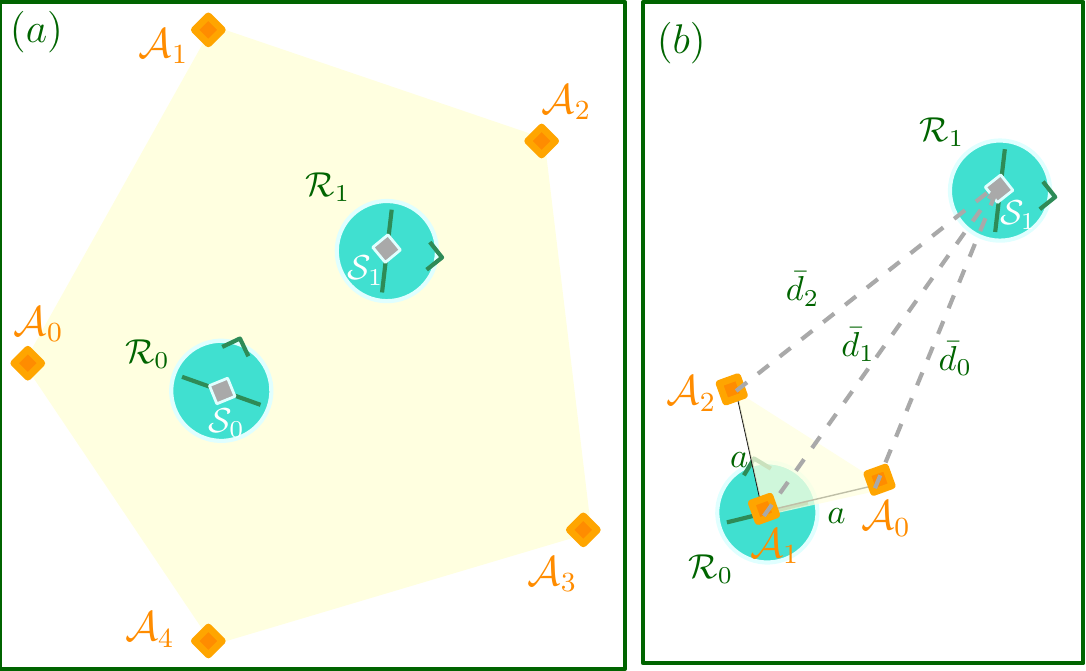}
	\caption{(a) A conventional localization framework and (b) an on-board localization framework}
	\label{fig:convexHull}
\end{figure}

The conventional mobile robot localization methods offer two ways to obtain the sensor data. In the first way, the robot position is obtained from a fixed infrastructure in a well-designed environment. A conventional setup for this approach comprises at least three anchors, a ground station, and sensors mounted on robots (Fig.~\ref{fig:convexHull}-a). The anchors are installed in a room at certain positions separated at the utmost distances from each other so that they form a large convex hull. Therefore, the mobile robot always moves inside the convex hull of the anchors. The ground station estimates the positions of the mobile robots and transmits the estimates to the robots continuously. Generally, this method yields highly accurate location estimate with high data rate. However, this framework entirely depends on the environment: The localization can be performed only in that particular environment. In the second way, the robot implements a localization algorithm with its on-board sensing and computational capabilities solely. The robot either measures its distances and bearing angles to specific landmarks or employs vision sensors to identify its location in a given map of the environment.

Recently, several works have combined the good sides of the two approaches for multi-robot systems under the framework of on-board anchor configuration (Fig.~\ref{fig:convexHull}-b) \cite{UwbEth_16,OnboardAnchor_Rus17,GulerCCTA}. In this framework, a robot is equipped with a set of anchors on-board, with the inter-anchor distances limited by the physical characteristics of the robot. Notably, the resulting anchor configuration forms a smaller convex hull than in the conventional anchor configuration case. This anchor robot localizes another robot that has a sensor on board. Therefore, the localized robot always lies outside of the anchors' convex hull. In \cite{UwbEth_16,OnboardAnchor_Rus17,GulerCCTA}, the filtering approaches are employed because the localization algorithm must deal with the sensor uncertainty in real time as well.

Here, we are interested in a real-time, completely distributed, infrastructure-free, and on-board localization algorithm for multi-robot systems. We propose a localization framework for a two-robot system that utilizes three UWB anchors on one robot and a single UWB sensor on the other. We use a dual Monte-Carlo localization (MCL) algorithm based on a particle filter to estimate the relative position between the two robots. The filter utilizes the most recent sensory information in the prediction phase and then evaluates the weights based on the assumed motion model. This framework allows the anchor robot to track the agile maneuvers of the localized robot with a small number of particles, which would require a larger number of particles with the conventional particle filtering. Remarkably, this framework does not utilize an explicit communication structure, i.e., the robots do not communicate with each other or with a ground station. Furthermore, we demonstrate with experiments that the estimation accuracy suffices to implement some formation control objectives such as relative position maintenance when integrated with simple motion control algorithms. Our contributions are as follows:
\begin{itemize}
	\item We propose an MCL approach based on an onboard UWB configuration for the multi-robot localization problem that is suited to handle agile robot motions by using a small number of particles.
	\item We relax the assumption of \textit{a priori} knowledge of the velocity profile of the localized robot, without imposing a communication framework.
	\item We combine the localization algorithm with simple motion control laws to solve some formation control objectives and demonstrate its experimental performance on a two-drone system.
\end{itemize}

The rest of the paper is organized as follows. Section~\ref{sec:relatedWorks} reviews the literature on the mobile robot localization. Section~\ref{sec:problemDef} gives the multi-robot localization problem in general terms. Section~\ref{sec:particleReview} summarizes the conventional and dual MCL algorithms. Section~\ref{sec:Framework2D} presents the proposed localization algorithm. Section~\ref{sec:simulations} and \ref{sec:experiments} demonstrate the simulation and experimental results. Section~\ref{sec:discussion} gives a discussion on the results. Finally, Section~\ref{sec:conclusion} contains concluding remarks.

\setcounter{equation}{0}
\section{Related Works}\label{sec:relatedWorks}
Mobile robot localization has been studied extensively in the literature, see  \cite{AutonomousMobile_04,ProbabilisticRob_05,PrincRobotMot_05,MFbook,Gezici_05,Thrun_RobustMCL01} and the references therein for a detailed survey. In the literature, the localization problem for a multi-robot system has been translated in two ways based on the control objectives: $(i)$ Each robot is localized in a global frame independently by a global localization system; $(ii)$ Each robot estimates its relative positions to other robots or objects in its own frame. Notably, $(i)$ is a straightforward extension of the single robot localization problem to the multi-robot case.

The majority of previous works on indoor localization utilized the conventional anchor configuration \cite{UWB_Singapore17,UwbEpfl_11,ParticleUwb_09} to solve $(i)$. Reference \cite{UWB_Singapore17} combines UWB sensors with a visual inertial system to correct drifts when building maps. The authors in \cite{ParticleUwb_09} utilize particle filtering to handle the multimodal error behavior of the non-line-of-sight (NLOS) UWB measurements. Furthermore, the conventional configuration is used in \cite{UwbEpfl_11} to solve the single-robot as well as the multi-robot localization problem. The authors in \cite{CoopLocal_Solmaz_arxiv,Kia2016Cooperative-Loc,Dudek_CoopLoc01,Particle_Prorok13} exploited the inter-robot communication and proposed cooperative EKF localization architectures to improve the estimation accuracy in multi-robot systems. The key idea in these works is that each robot receives estimation related information from its neighbor robots through communication. In outdoor environments, several works employed GPS sensors on board the robots to achieve formation control tasks. For instance, reference \cite{OutdoorSwarmGPS14} equipped every robot in a swarm with a GPS receiver and demonstrated a flock behavior with drones by the aid of inter-robot communication. Although the frameworks in \cite{CoopLocal_Solmaz_arxiv,Kia2016Cooperative-Loc,Dudek_CoopLoc01,OutdoorSwarmGPS14} improve the estimation performance for multi-robot systems with undirected graphs, they still depend on a global positioning system (GPS or Mocap) and bring an extra cost for the communication layer, which makes their reliability aspect questionable.

Toward the goal of freeing the localization framework from environment completely, recently several works considered an onboard anchor configuration where a moving vehicle equipped with a set of anchors localizes another robot or human \cite{UwbEth_16,UWBonboardNumerical_17,OnboardAnchor_Rus17,GulerCCTA}. In \cite{UwbEth_16}, a quadrotor equipped with UWB anchors on board tracks a target with a single UWB sensor by employing an iterated EKF. However, the approach of \cite{UwbEth_16} still depends on infrastructure because the quadrotor control relies on Mocap or GPS data instead of the localization feedback. In \cite{OnboardAnchor_Rus17}, a quadrotor moves in front of a ground vehicle and searches for safe paths for the vehicle while localizing itself with respect to the vehicle by unscented Kalman filter and optimization techniques. In \cite{GulerCCTA}, a mobile robot maintains its relative position to another robot by using feedback from a unique UWB localization framework. The main differences between the current work and \cite{GulerCCTA} are twofold. First, we propose an MCL algorithm here whereas \cite{GulerCCTA} proposed EKF based algorithms. Second, here we consider the VTOL vehicles particularly while non-holonomic ground robots were considered in \cite{GulerCCTA}.

Trilateration method takes a set of anchor-sensor distance measurements at a particular time instant and produces a closed-form solution for possible sensor locations based on the distance geometry \cite{Trilat_Algebraic96,Trilat_Geometric05,Trilateration_08}. Similarly, the optimization approach minimizes the additive noise on a set of anchor-sensor distances, subject to equalities obtained from the geometric properties \cite{OnboardAnchor_Rus17,CaoAndMorse_SCL06}. However, these two methods have not been preferred for mobile robot localization because they greatly suffer from measurement noises. To obtain reliable estimation results for mobile robots under noisy distance measurements, Bayesian methods were commonly employed. The authors in \cite{UltrasLocalzn_13} designed an EKF algorithm with sonar anchors. Reference \cite{UwbImuFusionETH_15} fused inertial and UWB sensors' data to estimate a quadrotor's position. In \cite{ParticleUwb_09}, a particle filter based localization algorithm was applied on UWB distance data for both LOS and NLOS measurement cases. Similarly, in \cite{UwbEpfl_11}, particle filtering was applied to localize single- and multi-robot systems in a well-designed environment. The authors in \cite{Particle_Prorok13} proposed a simple model that captures the multimodal error behavior of the UWB measurements in NLOS environments and designed a particle filter based localization algorithm for multi-robot systems in an indoor environment. Our framework differs from \cite{Trilat_Algebraic96,Trilat_Geometric05,Trilateration_08,OnboardAnchor_Rus17,CaoAndMorse_SCL06,UltrasLocalzn_13,UwbImuFusionETH_15,ParticleUwb_09,Particle_Prorok13,UwbEpfl_11} in that they utilized a set of UWB beacons located at known positions in a room to provide the robots with distance data, which makes the algorithms infrastructure-dependent.

The MCL algorithms can track the states of non-linear, non-Gaussian models \cite{Particle_tutorial02,ProbabilisticRob_05}. Particle filters were employed in the robot localization and mapping problems \cite{ParticleUwb_09,Particle_Gustafsson02,Doucet2000,Particle_Prorok13}, for object detection in images \cite{Particle_Detection03,Particle_Detection11}, in wireless communication \cite{Particle_Magazine03}, and in many other applications. The dual MCL approach was initially proposed in \cite{Thrun_RobustMCL01} to handle the particle depletion issue in cases where the state transition distribution covariance is incomparably higher than the measurement covariance. This technique was used in \cite{Burgard_DualMCL07} to solve the grid mapping problem with precise laser range finders. This framework, with suitable modifications, fits well our particular problem setting because the UWB measurements produce a better prediction about an agile robot's current location than the state transition distribution of the robots.

\setcounter{equation}{0}
\section{System Definition}\label{sec:problemDef}
Consider a two-robot system $\Msf=\{\Rz,\Ro\}$, where $\Ri$ denotes the $i$th robot. We focus on the two-dimensional Euclidean plane case as the configuration space of $\Msf$. We consider a holonomic kinematics model in discrete-time for each robot $\Ri$ as follows:
\begin{align}
\begin{bmatrix}
\prm^{i}_{k+1}\\
\vrm^{i}_{k+1}
\end{bmatrix}
&=
\begin{bmatrix}
\Irm_{2} & T_{s}\Irm_{2}\\
0_{2} & \Irm_{2}
\end{bmatrix}
\begin{bmatrix}
\prm^{i}_{k}\\
\vrm^{i}_{k}
\end{bmatrix}
+
\begin{bmatrix}
(T_{s}^2/2)\Irm_{2}\\
T_{s}\Irm_{2}
\end{bmatrix}
\arm^{i}_{k}
+
\begin{bmatrix}
0_{2}\\
\delta^{i}_{k}
\end{bmatrix},\label{eq:holModel}
\end{align}
where $\prm^{i}=\left[x^{i},y^{i}\right]^{\top}\in\Re^{2}$ is the position, $\vrm^{i}\in\Re^{2}$ is the velocity, and $\arm^{i}\in\Re^{2}$ is the acceleration of robot $\Ri~(i=0,1)$, $k$ is the time step, $T_{s}$ is the sampling time, and $\delta^{i}$ is the random-walk process noise with the following profile:
\begin{align}
\delta^{i}_{k}\sim\Ncal(0_{2},\Qrm^{\irm}_{\textrm{mot}}),\label{eq:motNoiseModel}
\end{align}
with $\Qrm^{\irm}_{\textrm{mot}}\in\Re^{2\times2}$ being the noise covariance matrix. We assume that the velocities and accelerations of the robots are saturated by their maximum values as follows:
\begin{align*}
\mathrm{v^{i}_{\min}}&\leq\vrm^{i}_{k}\leq\mathrm{v^{i}_{\max}},\quad \mathrm{a^{i}_{\min}}\leq\arm^{i}_{k}\leq\mathrm{a^{i}_{\max}},\quad
i=\{0,1\},
\end{align*}
where the constant vectors $\mathrm{v^{i}_{\min}},\mathrm{v^{i}_{\max}},\mathrm{a^{i}_{\min}},\mathrm{a^{i}_{\max}}\in\Re^{2}$ are pre-defined based on the physical characteristics of the robots.
\begin{remark}\label{rem:motionModel}
The model \eqref{eq:holModel} approximates the behavior of a second-order mechanical system where the acceleration between two successive time steps is assumed constant. For instance, a quadrotor obeys this motion behavior when its internal nonlinear dynamics are controlled by a low-level, high-frequency microcontroller. 
\end{remark}

Let $\Fcal_{\Grm}\subseteq\Re^{2}$ denote the global frame, and let $\Fcal_{\mathrm{i}}$ denote the body frame of $\Ri$. We denote robot $\Rz$ as the \textit{anchor robot} and mount three UWB anchors on robot $\Rz$ at positions $\qrm_{1},\qrm_{2},\qrm_{3}$ on the frame $\Fcal_{0}$ as follows (Fig.~\ref{fig:convexHull}-b):
\begin{align}
\qrm_{1}&=\left[a,0\right]^{\top},~\qrm_{2}=\left[0,0\right]^{\top},~\qrm_{3}=\left[0,a\right]^{\top},\label{eq:anchorPose}
\end{align}
where the design parameter $a$ is determined based on the physical characteristics of robot $\Rz$. Thus, the three anchors are rigidly linked to each other on $\Rz$. We set $a=1$ in the algorithm design process and discuss the modifications for different values of $a$ in Section~\ref{sec:experiments}.

We mount a UWB sensor at $\prm_{1}$, the center of frame $\Fcal_{1}$. As anticipated from the geometry of the system, $\prm_{1}$ always remains outside of the convex hull of the three points $(\qrm_{1},\qrm_{2},\qrm_{3})$. We denote the distances between each $\qrm_{i}$ and $\prm^{1}$ by $\bar{d}^{i}$:
\begin{align}\label{eq:distDefCorrect}
\bar{d}^{i}_{k}&=\|\prm^{1}_{k}-\qrm_{i}\|,\quad i\in\{1,2,3\}.
\end{align}
We assume that the distance measurements are collected at the UWB anchors on robot $\Rz$ and corrupted by additive noise $\eta^{i}$ such that $d^{i}_{k}=\bar{d}^{i}_{k}+\eta^{i}_{k}.$ We consider the following state vector:
\begin{align}\label{eq:stateVector}
\xrm_{k}&=\left[
\rrm_{k},~
\mathrm{v^{1g}_{k}}\right]^{\top},
\end{align}
where
\begin{align}\label{eq:relPosDef}
\rrm_{k}&=\prm^{1}_{k}-\prm^{0}_{k}
\end{align}
is the relative position between robot $\Rz$ and $\Ro$ on frame $\Fcal_{0}$, and $\mathrm{v^{1g}}$ is the velocity of robot $\Ro$ in frame $\Fcal_{G}$. We denote the estimate of $\xrm_{k}$ by $\hat{\xrm}_{k}$.

Thus, the dynamics of the relative position $\rrm$ is trivially derived as follows:
\begin{align}\label{eq:relPosDynAct}
\rrm_{k+1}&=\rrm_{k}+(\vrm^{1}_{k}-\vrm^{0}_{k})T_{s}+0.5(\arm^{1}_{k}-\arm^{0}_{k})T_{s}^{2}.
\end{align}

We assume that the robots do not have an explicit communication structure, i.e., they do not exchange information between each other. Furthermore, we assume that we do not have a ground station that can collect sensory information or implement the estimation algorithm. Therefore, the robots utilize their on-board sensors solely for the estimation and motion control objectives.

Robot $\Rz$ knows that robot $\Ro$ moves based on the motion model \eqref{eq:holModel}. Robot $\Rz$ also knows the maximum speed $\mathrm{v^{1}_{max}}$ of robot $\Ro$. However, robot $\Rz$ does not have access to the instant velocity $\mathrm{v^{1,G}_{k}}$. This assumption reflects a realistic multi-robot scenario where each robot is informed with the motion capabilities of the other robot but cannot access the instant values of the other robot's states such as position and velocity.

We define the objective in this paper as follows. Given the system $\Msf$ of robots $\Rz,\Ro$ with the motion model \eqref{eq:holModel}, the noisy distances $d^{1}_{k},d^{2}_{k},d^{3}_{k}~(k\geq1),$ a suitable initial condition $\hat{\xrm}_{0}$, and the aforementioned assumptions on the communication structure, generate the estimate $\hat{\xrm}_{k}$ of $\xrm_{k}$ so that the error 
\begin{align*}
e_{k}=\|\hat{\xrm}_{k}-\xrm_{k}\|
\end{align*}
is minimized for $k\geq1$.

\setcounter{equation}{0}
\section{Particle Filter Review}\label{sec:particleReview}
As a non-parametric Bayes filter, a particle filter does not use a compact model to represent the state distribution, unlike the Kalman filter and its variations. Instead, it uses a large number of samples to represent the current belief about the state. At any time, each sample denotes the algorithm's hypothesis on where the system state may lie. Similar to other Bayesian filters, a particle filter generates the state estimate in two phases: the prediction phase and the update phase. The resampling process in the update phase forms an important part of particle filters, where the samples are rearranged based on the current exteroceptive measurement data. Particle filters can be applied to models where the noise shows a non-Gaussian behavior because particle filters can inherently track any distribution under certain assumptions. We now summarize the conventional particle filter algorithm and refer the reader to \cite{Particle_Magazine03,Particle_Gustafsson02,Particle_Gustafsson10} for detailed descriptions.

We denote by $\mathrm{bel}(\xk)$ the current belief, or the posterior probability, of the state distribution, given as follows:
\begin{align}\label{eq:beliefDefBayes}
\mathrm{bel}(\xk)&= \pi(\xk|\zrm_{1:k},\urm_{1:k}),
\end{align}
where $\urm_{1:k}$ and $\zrm_{1:k}$ denote the inputs and measurements up to time step $k$, respectively. At any time step $k$, where $t=kT_{s}$ with $T_{s}$ being the sample time, a particle filter approximates the distribution \eqref{eq:beliefDefBayes} with the set of samples
\begin{align}\label{eq:sampleSet} \mathrm{S}_{k}=\lbrace\mathrm{s}^{1}_{k},\cdots,\mathrm{s}^{m}_{k}\rbrace,
\end{align}
where $m\in\mathbb{Z}_{+}$ denotes the number of samples, and $\mathrm{s}^{i}_{k}=\lbrace \mathrm{x}^{i}_{k},w^{i}_{k}\rbrace$ denotes the $i$th sample with the state $\mathrm{x}^{i}$ and the importance weight $w^{i}\in[0,1)$. Each $\mathrm{x}^{i}_{k}$ represents the hypothesis on the actual state value $\mathrm{x}_{k}$. Therefore, a particle filter represents the posterior state distribution by the discrete set $\mathrm{S}_{k}$, that is,
\begin{align}\label{eq:beliefDef}
\mathrm{bel}(\mathrm{x}_{k})&\sim \mathrm{S}_{k}.
\end{align}

The belief $\mathrm{bel}(\mathrm{x}_{k})$ in \eqref{eq:beliefDef} represents an approximation of the posterior distribution in \eqref{eq:beliefDefBayes} at any time step $k$ with $m$ number of samples $\mathrm{s}^{i}$. Admittedly, as $m$ approaches infinity, the belief improves, i.e., $\mathrm{bel}(\mathrm{x}_{k})$ approaches to $\pi(\mathrm{x}_{k}|z_{1:k},u_{1:k})$, at the expanse of increased computational complexity. Usually, $m$ is chosen large, e.g., $m>1000$. In the prediction phase, a new set of hypotheses are constructed from the previous sample set $\mathrm{S}_{k-1}$ based on the proposal distribution
\begin{align}\label{eq:proposalDistribution}
\varphi_{k}&=\pi(\xrm^{i}_{k}~|~\urm_{1:k},\zrm_{1:k})\mathrm{bel}(\xrm^{i}_{k-1}).
\end{align}
Subsequently, in the update phase, the weights $w^{i}_{k}$ are calculated as the fraction of the target distribution to the proposal distribution. Finally, the samples are rearranged with respect to their weights in the resampling process.

Although the particle filter process is determined by \eqref{eq:proposalDistribution} and the aforementioned weight calculation method, a designer can adopt different proposal distributions and resampling processes for a particular design. A common practice is to use the robot motion model as the proposal distribution to propagate the samples, as follows:
\begin{align}\label{eq:beliefPrediction}
\varphi_{k}&\sim \pi(\xrm_{k}|\xrm_{k-1},\urm_{k}),
\end{align}
where $\pi(\xrm_{k}|\urm_{k},\xrm_{k-1})$ is the robot's state transition distribution which depends only on the last state and the current input. Accordingly, the weights are calculated as
\begin{align}\label{eq:weightUpdate}
w^{i}_{k}&=\alpha\pi(z_{k}|x^{i}_{k}),
\end{align}
which corresponds to the observation model. Therefore, the new set of samples $\mathrm{S}_{k}$ constructed with a resampling process represents the posterior probability $\mathrm{bel}(\mathrm{x}_{k})$.

Although the framework \eqref{eq:beliefPrediction}-\eqref{eq:weightUpdate} generally yields high performance, it may cause the particle depletion problem in some applications including our specific problem. Especially, when the variance of the exteroceptive measurement model is much lower than the variance of the robot's state transition distribution, propagating the particles with the proposal distribution \eqref{eq:beliefPrediction} may populate most of the particles in regions that do not align with the exteroceptive measurement model. This misalignment would set the weights of the majority (or all) of the particles to small values and reduce the efficiency of the resampling process. To address this issue, several alternative proposal distributions have been proposed. In \cite{Blanco_Optimal10,Doucet2000,Thrun_RobustMCL01}, the authors invert the roles of the prediction and update phases. In \cite{Thrun_RobustMCL01}, the measurement model is used in the proposal distribution:
\begin{align}\label{eq:alternativeProposal}
\varphi_{k}&=\frac{\pi(z_{k}~|~x_{k})}{\pi^{n}_{k}},
\end{align}
where $\pi^{n}$ is a normalizer. Accordingly, they use the following importance weights:
\begin{align}\label{eq:alternativeWeightUpdate}
w^{i}_{k}&=\pi(\xrm_{k}|\urm_{1:k},\zrm_{1:k-1}),
\end{align}
where $\pi(\xrm_{k}~|~\urm_{1:k},\zrm_{1:k-1})$ is calculated by an extra sampling process at each time step. In other words, the belief is predicted with the exteroceptive measurements, and the update is performed based on the motion model, in contrast to the conventional algorithm. Therefore, this approach populates the particles around the most recent observation and hence solves the particle depletion issue for some scenarios. We adopt this approach to solve our localization problem in the following section.

\setcounter{equation}{0}
\section{The Proposed Framework} \label{sec:Framework2D}
We present our localization algorithm in this section. We aim to design a distributed algorithm where robot $\Rz$ estimates the relative position $\mathrm{r}$ to robot $\Ro$ in its local frame $\Fcal_{0}$ by using only its own computational devices. We assume that no central computational unit (e.g. a ground station) exists. The robots do not explicitly communicate with each other. However, the robots sense ranges with the on-board UWB sensors by an implicit communication mechanism, which we consider as a ranging mechanism similar to the case of the laser range finder with a receiver. The proposed algorithm should provide the position estimate in real-time by using the on-board sensors of robot $\Rz$. Since the measurement data acquired from the sensors are noisy, the algorithm has to deal with uncertainties. Also, the performance of the proposed algorithm should suffice to be used as a feedback to further motion control algorithms on robot $\Rz$.

In the remainder of this section, we propose our localization framework for a two-robot system. We describe the details of the algorithm in section~\ref{sec:blockDiag}-\ref{sec:resampling}. Then, we give the pseudo-code of our algorithm in \ref{sec:algorithmComplexity}. We give the calibration procedure of the UWB sensors in Section~\ref{sec:UwbCharacterization}. We discuss the details of data acquisition and implementation in Section~\ref{sec:experiments}.

\begin{figure}[!ht]
	\centering
	\includegraphics[trim = 0cm 0cm 0cm 0cm, clip, width=0.35\textwidth]{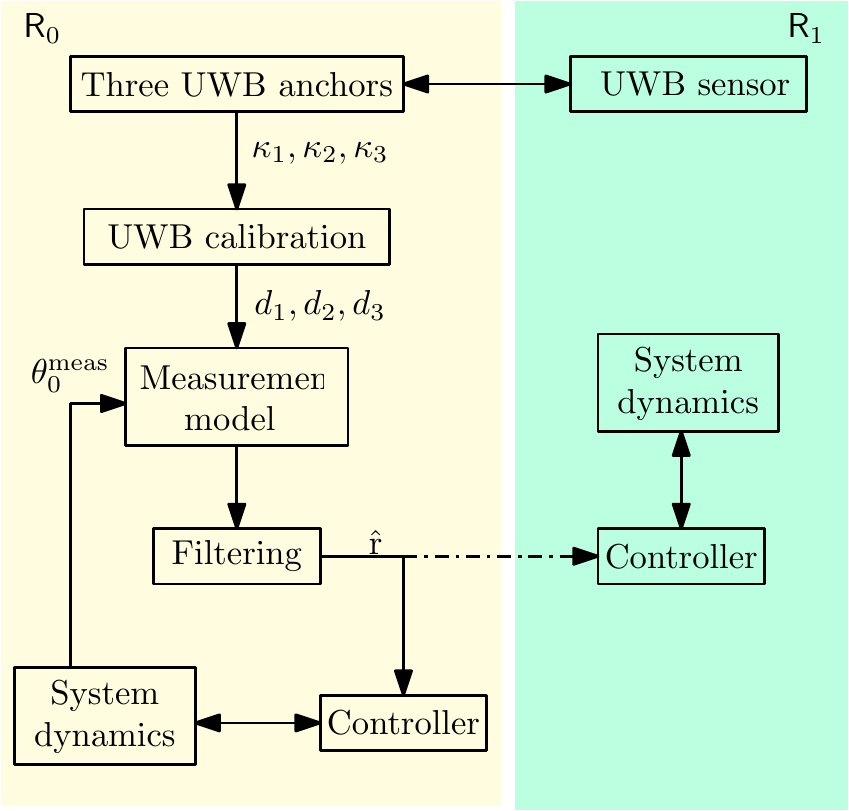}
	\caption{The block diagram of the proposed framework. Robots $\Rz$ and $\Ro$ are represented with yellow and green backgrounds, respectively. The dashed arrow shows the optional data transmission from $\Rz$ to $\Ro$.}
	\label{fig:block}
\end{figure}

\subsection{The Block Diagram}\label{sec:blockDiag}
The block diagram of the two robot system $\Msf$ is depicted in Fig.~\ref{fig:block}. Robot $\Rsf_{0}$ is to generate the estimate vector $\hat{\mathrm{x}}$. The filtering algorithm runs on board the anchor robot $\Rz$. Three raw UWB distance values are acquired from the UWB anchors and passed to the \textsc{UWB calibration} block to process the distance data and eliminate possible biases based on a calibration procedure. Afterwards, the \textsc{Measurement model} block accepts the three distance measurements and the IMU measurement of $\Rsf_{0}$, and outputs the ``constructed'' measurements. Finally, the \textsc{Filtering} block generates the state estimate $\hat{\xrm}$, which is then relayed back to the motion controller of $\Rsf_{0}$ to close the motion control loop. Also, the state estimate can be transmitted to robot $\Rsf_{1}$ by communication to allow $\Rsf_{1}$ to use the estimate for better formation control performance, but we do not consider that case in this paper. We assume that the estimation takes place and is used in robot $\Rsf_{0}$ solely. 

In the remainder of this section, we propose the localization algorithm by assuming that the motion of each robot is controlled by its own low-level motion controller which is commanded by exogenous inputs. We study the integration of the localization output with the motion control algorithms in Section~\ref{sec:simulations} and \ref{sec:experiments}.

\subsection{The Proposal Distribution}\label{sec:proposedAlgo}
The authors in \cite{GulerCCTA,UwbEth_16} employed EKF to estimate the relative position between two robots by assuming that the velocity of robot $\Rsf_{1}$ is unknown but either constant or slightly varying. Likewise, here we assume that robot $\Rsf_{0}$ does not have access to the instant velocity of robot $\Rsf_{1}$. Robot $\Rsf_{1}$ can be a slowly moving ground robot or an aerial vehicle with agile motion behavior. Our design aims at yielding good estimation performance for a broad spectrum of motion characteristics for robot $\Rsf_{1}$ including aggressive maneuvers. We exploit the non-parametric nature of particle filters to estimate $\rrm$ for different motion behaviors of robot $\Rsf_{1}$.
\begin{figure}[!b]
	\centering
	\includegraphics[trim = 6.0cm 0.0cm 6.0cm 5cm, clip, width=0.35\textwidth]{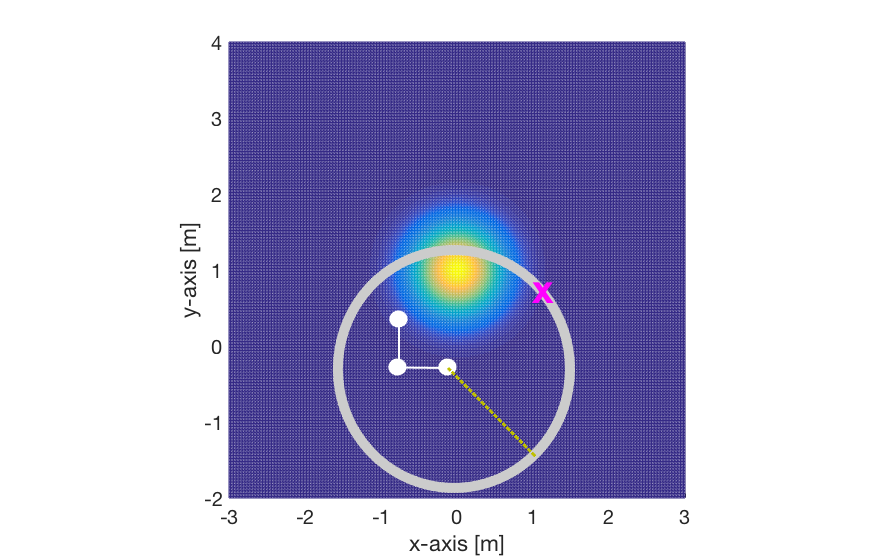}
	\caption{The likelihood of the motion model (demonstrated as a Gaussian distribution originated at $\prm=[0,1]^{\top}$m), the likelihood of the first sensor's observation model (demonstrated as the gray circle), three onboard UWB anchors (demonstrated as white dots), and the true location of robot $\Rsf_{1}$ (demonstrated as the magenta cross). If only the motion model is used for the proposal distribution, the particles would condense at the peak of the Gaussian distribution (shown in yellow) and likely miss the robot $\Rsf_{1}$'s true location.}
	\label{fig:proposal_dist}
\end{figure}

We argue that the common practice, i.e. using the robot motion model as the proposal distribution as in Section~\ref{sec:particleReview}, can yield poor performance for our particular problem, especially when a small number of particles are used. In Fig.~\ref{fig:proposal_dist}, we illustrate a reason why the estimation performance may degrade with such a proposal distribution. Consider the problem definition in Section~\ref{sec:problemDef}. Assume that robot $\Rsf_{1}$ moves with a slightly varying velocity $\vrm^{1}$. If the conventional algorithm is used, then the particles are propagated based on the motion model (which is assumed Gaussian in Fig.~\ref{fig:proposal_dist}), and the majority of the particles will be condensed at the peak region of the state transition distribution at the end of the prediction phase (depicted in yellow--orange). However, robot $\Rsf_{1}$ may drift away from its estimated region between the two successive time steps due to a disturbance or it might be that it moves fast in contrast to what was assumed. For instance, it might be located at the red circle in Fig.~\ref{fig:proposal_dist}. In such a case, only few particles would survive in the resampling process in which the distance measurements are evaluated. The likelihood region for one distance measurement is demonstrated in gray in Fig.~\ref{fig:proposal_dist}. The repetition of this process would likely cause the particle depletion issue. A solution to the particle depletion requires to use a large number of particles, e.g., more than $5000$, which is computationally inefficient. 

\begin{figure}[!t]
	\centering
	\includegraphics[trim = 2.8cm 7.2cm 4.5cm 2cm, clip, width=0.35\textwidth]{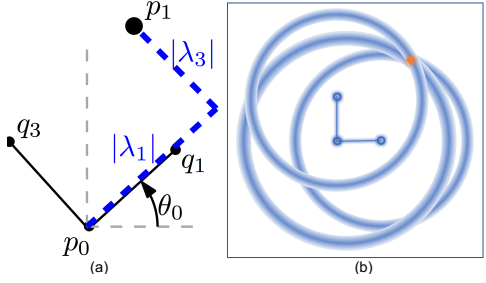}
	\caption{(a) Construction of $\mathrm{r^{meas}}$; (b) A representation of the uncertainty model of $\mathrm{r_{meas}}$. The three circles depict the uncertainty regions of the distance measurements within a certain $\sigma$ bound (the darker the color the more likely the target may be). The target is shown in orange.}
	\label{fig:relpos_distuncert}
\end{figure}

Inspired by the dual MCL approach of \cite{Thrun_RobustMCL01}, we now design the proposal distribution and resampling process of our particle filter. We use the proposal distribution \eqref{eq:alternativeProposal} to propagate the particles. To model the distribution $\pi(z_{k}|x_{k})$, we note that the exteroceptive measurement model consists of three independent UWB measurements $d^{1}_{k},d^{2}_{k},d^{3}_{k}$. We use the constructed measurement model proposed in \cite{GulerCCTA} to map the distance measurements to an estimate location on the $x$-$y$ plane by using virtual axes as follows:
\begin{align}
\mathrm{r^{meas}_{k}}&=\lambda^{1}_{k}\qrm_{1}+\lambda^{3}_{k}\qrm_{3}\label{eq:barycenCoord}\\
&=\left[\lambda^{1}_{k},\lambda^{3}_{k}\right]^{\top},
\end{align}
where $\mathrm{r^{meas}}$ is the position of $\Rsf_{1}$ in frame $\Fcal_{0}$, $\qrm_{1},\qrm_{3}$ are the anchor locations, and $\lambda^{i}=s^{i}|\lambda^{i}|$ with $s^{i}=\text{sgn}(\lambda^{i})$ are the coordinates of $p_{1}$ in $\Fcal_{0}$ (Fig.~\ref{fig:relpos_distuncert}-a). Notably, the line segments $l(\qrm_{2},\qrm_{1}),l(\qrm_{2},\qrm_{3})$ form the virtual $x$-$y$ axes of $\Fcal_{0}$. We have the following geometric relations \cite{Barycentric,CaoAndMorse_SCL06}:
\begin{align}
|\lambda^{1}_{k}|&=\dfrac{|\Acal(\prm^{1}_{k},\qrm_{2}[k],\qrm_{3}[k])|}{|\Acal(\qrm_{1},\qrm_{2},\qrm_{3})|},\label{eq:barycenArea1}\\ |\lambda^{3}_{k}|&=\dfrac{|\Acal(\prm^{1}_{k},\qrm_{1}[k],\qrm_{2}[k])|}{|\Acal(\qrm_{1},\qrm_{2},\qrm_{3})|}\label{eq:barycenArea2},\\
s^{i}&=\text{sgn}(d_{2}^2+1-d_{i}^2),
\end{align}
where $\Acal(\qrm_{1},\qrm_{2},\qrm_{3})=0.5a^{2}$ denotes the area of the right triangle formed by the three anchors on robot $\Rsf_{0}$. Without loss of generality, assume $a=1$m. Then, it follows that \cite{Barycentric,CaoAndMorse_SCL06}
\begin{align*}
|\Acal(\prm_{1},\qrm_{2},\qrm_{3})|&=\dfrac{1}{4}
\sqrt{\left(d_{2}^2-d_{3}^2\right)^{2} - 2\left(d_{2}^2+d_{3}^2\right)+1},\\
|\Acal(\prm_{1},\qrm_{1},\qrm_{2})|&=\dfrac{1}{4}
\sqrt{\left(d_{1}^2-d_{2}^2\right)^{2} - 2\left(d_{1}^2+d_{2}^2\right)+1}.
\end{align*}

\begin{remark}
The magnitudes of $\lambda^{1},\lambda^{3}$ are well defined in \eqref{eq:barycenArea1},\eqref{eq:barycenArea2} because the terms $\Acal(\prm^{1}[k],\qrm_{2}[k],\qrm_{3}[k]),~\Acal(\prm^{1}[k],\qrm_{1}[k],\qrm_{2}[k])$ are defined when either of the vertex sets $\{\prm^{1},\qrm_{2},\qrm_{3}\},~\{\prm^{1},\qrm_{1},\qrm_{2}\}$ are collinear and thus does not form a triangle.
\end{remark}

We now calculate the distribution $\pi(z_{k}|x_{k})$ by using the constructed vector $\mathrm{r^{meas}_{k}}$. If the objective was to localize the robot in a given map, then a common method to construct $\pi(z_{k}|x_{k})$ is to take a large number of sensor measurements, build the joint distribution $\pi(z_{k},x_{k})$, and form a grid map by kd-trees conditioned on some functions of collected features \cite{Thrun_RobustMCL01}. Since our problem statement does not include a map, we use a direct approach to obtain the distribution $\pi(z_{k}|x_{k})$.

Since the construction of $\mathrm{r^{meas}}$ involves arithmetic operations, the uncertainty model of $\mathrm{r^{meas}}$ greatly differs from the uncertainty model of the distance measurements $d^{i}$. In Section~\ref{sec:UwbCharacterization}, we model the uncertainty characteristics of the distance measurements $d^{i}$. Assuming that the bias term $b_{i}$ is perfectly compensated, the additive noise on the distance measurements can be modeled by a Gaussian distribution with zero mean and a variance which is specific to the anchor. Each distance measurement produces a circular likelihood region for the relative position $\rrm_{k}$. Remarkably, in the absence of noise, the three distance measurements intersect at the true relative position $\rrm_{k}$ (Fig.~\ref{fig:relpos_distuncert}-b). Therefore, the resulting configuration attains the highest probability at $\rrm_{k}$ and less probability around $\rrm_{k}$. We approximate this uncertainty model as a Gaussian distribution centered at $\mathrm{r^{meas}}$ as follows:
\begin{align}\label{eq:measUncertaintyApprox}
\pi(z_{k}|x_{k})&\approx\Ncal\left(\mathrm{r^{meas}_{k}},\Qrm_{\mathrm{obs}}\right),
\end{align}
where $\Qrm_{\mathrm{obs}}=\diag\left(\sigma_{\mathrm{x,obs}}^{2},\sigma_{\mathrm{y,obs}}^{2}\right)$ denotes the measurement covariance matrix and is a design parameter, and $\sigma_{\xrm,\mathrm{obs}}$ and $\sigma_{\yrm,\mathrm{obs}}$ denote the standard deviations in the $x$ and $y$ axes in frame $\Fcal_{0}$, respectively. Accordingly, we sample the particles based on the proposal distribution:
\begin{align}\label{eq:sample}
x^{i}_{k}&\sim\varphi_{k},\quad i=(1\ldots m),
\end{align}
where $\varphi_{k}$ is as in \eqref{eq:alternativeProposal} with $\pi(z_{k}|x_{k})$ as defined in \eqref{eq:measUncertaintyApprox}. In summary, we generate $m$ particles around the constructed measurement vector $\mathrm{r^{meas}_{k}}$ to represent the state hypothesis.

The design parameters $\sigma_{\xrm,\mathrm{obs}},\sigma_{\yrm,\mathrm{obs}}$ can be found empirically with numerical simulations. Evidently, a set of distance measurements with high variances will result in high values for $\sigma_{\xrm,\mathrm{obs}}$ and $\sigma_{\yrm,\mathrm{obs}}$. We suggest to use the values that yield the best observed performance. Notably, high $\sigma_{\xrm,\mathrm{obs}},~\sigma_{\yrm,\mathrm{obs}}$ values would require a larger number of particles than small $\sigma_{\xrm,\mathrm{obs}},~\sigma_{\yrm,\mathrm{obs}}$ values. We emphasize that the approximations for these parameters are expected to perform well because the particle filter does not require a perfect measurement. The critical aspect is that we need to represent the true state values with a subset of particles to avoid the particle depletion problem. In our experiments, we obtained a sufficiently good performance with a small number of particles and with a set of parameter values found empirically.

\begin{remark}
One might also think of propagating the particles based on the three distance measurements directly. However, as can be seen in Fig.~\ref{fig:relpos_distuncert}-b, one would need a large number of particles to cover all the three circular areas formed by the uncertainty models of the individual distance measurements. That is, we would need to fill the entire circles instead of the region around $\mathrm{r^{meas}_{k}}$. Therefore, this method would not yield a computationally efficient algorithm.
\end{remark}

\subsection{Resampling based on Motion Model}\label{sec:resampling}
We now model the distribution $\pi(\xk|\zrm_{1:k-1},\urm_{1:k})$ to calculate the importance weights $w^{i}$. The distribution $\pi(\xk|\zrm_{1:k-1},\urm_{1:k})$ is analogous to the predicted belief distribution of Kalman filtering where the belief is calculated by processing the last action $\urm_{k}$ before utilizing the last observation $\zrm_{k}$. In \cite{Thrun_RobustMCL01}, the authors propose to use the kernel density estimation method to construct the distribution $\pi(\xk|\zrm_{1:k-1},\urm_{1:k})$. In this method, every particle in $\mathrm{bel}(\xrm_{k-1})$ is propagated through the motion model $\pi(\xrm_{k}|\urm_{k},\xrm_{k-1})$. The new particles construct the kd-tree which represents the likelihood of the particles based on the motion model. 

In our framework, the distribution $\pi(\xrm_{k}|\urm_{k},\xrm_{k-1})$ stands for the dynamics \eqref{eq:relPosDynAct}. We assumed that robot $\Rz$ does not have access to the instant velocity of robot $\Ro$ but has a rough knowledge about its state transition distribution. This uncertainty can be modeled with any distribution scheme including Gaussian distribution, multimodal Gaussian distribution, and beta distribution, based on the \emph{a priori} knowledge on the motion behavior of robot $\Ro$. Remarkably, the Gaussian and uniform distributions are good candidates to approximate the state transition of robots with agile maneuver capabilities. Since here we consider a broad spectrum of robot motion capabilities, we exploit this assumption and model the state transition distribution of robot $\Ro$ as a normal distribution centered at the previous estimate vector $\hat{\vrm}^{1}_{k-1}$. We generate $m$ estimates of the velocity vector as follows:
\begin{align}\label{eq:v1Pred}
\tilde{\vrm}^{i}_{k}&\sim\Ncal\left(\hat{\vrm}^{1}_{k-1},\Qrm_{\mathrm{1,mot}}\right),
\end{align}
where $i\in\{1,\ldots,m\}$ and
\begin{align}\label{eq:QmotModel}
\Qrm_{\mathrm{obs}}&=\diag\left(\sigma_{\xrm,\mathrm{mot}}^{2},\sigma_{\yrm,\mathrm{mot}}^{2}\right)
\end{align}
is a design parameter that can be chosen suitably based on the application. For instance, small values in the diagonal entries can be used for non-holonomic vehicles with slow angular velocities while relatively higher values can be used for holonomic vehicles with aggressive maneuvers. We now relate this estimate to the likelihood of the relative position estimate $\mathrm{\hat{r}}_{k}$. Our aim is to find the velocities $\tilde{\vrm}^{i}_{k}$ that yield relative position estimates $\mathrm{\hat{r}}$ that are close to the last measurement $\mathrm{r}_{k}^{\mathrm{meas}}$. For this purpose, we generate $m$ auxiliary vectors
\begin{align}\label{eq:v1PredAuxiliaryState}
\tilde{\rrm}^{i}_{k}&=\mathrm{r^{avg}}_{k}+\tilde{\vrm}^{i}_{k}T_{s},
\end{align}
and evaluate their distance to the predicted belief particles by incorporation $\tilde{\rrm}^{i}$ into $w_{k}^{i}$:
\begin{align}\label{eq:w_iProposed}
w^{i}_{k}\sim&\pi_{\mathrm{aux}}(\xrm^{i}_{k}|\zrm_{1:k-1},\urm_{1:k})=\pi(\mathrm{\hat{r}}^{i}_{k}|\urm_{k},\mathrm{\hat{r}}_{k-1})\pi(\tilde{\rrm}^{i}_{k}|\zrm_{k}).
\end{align}
In other words, we aim to assign high probabilities for $\mathrm{\hat{r}}^{i}_{k}$ and $\tilde{\rrm}^{i}_{k}$ that are closest to $\mathrm{r}_{k}^{\mathrm{meas}}$.

\subsection{The Algorithm}\label{sec:algorithmComplexity}
We give the proposed localization algorithm in Algorithm~\ref{alg:propAlgo}. Algorithm~\ref{alg:propAlgo} requires a two-robot system $\Msf$ with the on-board UWB sensor configuration given in Section~\ref{sec:problemDef}. At any time step $k$, the algorithm receives the previous particle set $\mathrm{S}_{k-1}$, the three distance measurements $d^{i}_{k}$, and the control input $\vrm_{0}^{k}$ of robot $\Rz$ and generates the new particle set $\mathrm{S}_{k}$. First, the raw estimate $\mathrm{r}_{k}^{\mathrm{meas}}$ is constructed from $d^{i}_{k}$ (line~\ref{alg_constructRmeas}). Then, a new set of particles are sampled around $\mathrm{r}_{k}^{\mathrm{meas}}$ (line~\ref{alg_sampleMeas}). Next, a kernel density $\pi_{\mathrm{aux}}(\xrm^{i}|\zrm_{1:k-1},\urm_{1:k})$ is calculated by propagating the particles in $\mathrm{S}_{k-1}$ with the motion model of the robots. This process is represented as the \textsc{FindDensity} function in line~\ref{alg_kd}. Afterward, the importance weights $w^{i}$ are calculated by evaluating the particles $\xrm^{i}$ with respect to $\pi_{\mathrm{aux}}$ (line~\ref{alg_importanceW}). Finally, the belief is updated by resampling the particle set with the importance weights $w^{i}$ (line~\ref{alg_resample}).

\begin{algorithm}[!htb]
\caption{\small{The Proposed Dual MCL Algorithm}}
\label{alg:propAlgo}
\begin{algorithmic}[1]
\small{
\Require $\chi_{k-1},\urm_{k},\zrm_{k}$
\Ensure $\chi_{k},\hat{\rrm}_{k}$
\State Initialize $k=0,~\bar{\chi}_{0}=\emptyset,~\chi_{0}\sim \pi(\xrm_{0})$
\While{True}
\State\label{alg_kIncr} $k\gets k+1$
\State\label{alg_constructRmeas} Calculate $\mathrm{r}_{k}^{\mathrm{meas}}$
\For{$i=1\ldots N$}\label{alg_sampleMeasFor}
\State\label{alg_sampleMeas} Generate $\xrm^{i}_{k}\sim\Ncal\left(\mathrm{r}_{k}^{\mathrm{meas}},\Qrm_{\mathrm{obs}}\right)$ for $i=(1,\ldots,m)$
\EndFor
\State\label{alg_kd} $\pi_{\mathrm{aux}}(\xrm_{k}|\zrm_{1:k-1},\urm_{1:k})\gets$ \Call{FindDensity}{$\mathrm{bel}(\xrm_{k-1}),\vrm^{0}_{k},\hat{\vrm}^{1}_{k-1},\Qrm_{\mathrm{mot}}$}
\For{$i=1\ldots N$}\label{alg_sampleMotFor}
\State\label{alg_importanceW} Calculate $w^{i}_{k}\sim\pi_{\mathrm{aux}}(\xrm^{i}_{k}|\zrm_{1:k-1},\urm_{1:k})$
\State\label{alg_setForming} $\bar{\chi}_{k}\gets\bar{\chi}_{k}+\lbrace \mathrm{x}^{i}_{k},w^{i}_{k}\rbrace$
\EndFor
\State\label{alg_resample} $\chi_{k}\gets$\Call{Resample}{$\bar{\chi}_{k}$}
\State $\bar{\chi}_{k}=\emptyset$
\EndWhile
}
\end{algorithmic}
\end{algorithm}

\setcounter{equation}{0}
\section{Simulations}\label{sec:simulations}
\begin{figure}[!b]
	\centering
	\includegraphics[trim = 0cm 0cm 0cm 0cm, clip, width=0.3\textwidth]{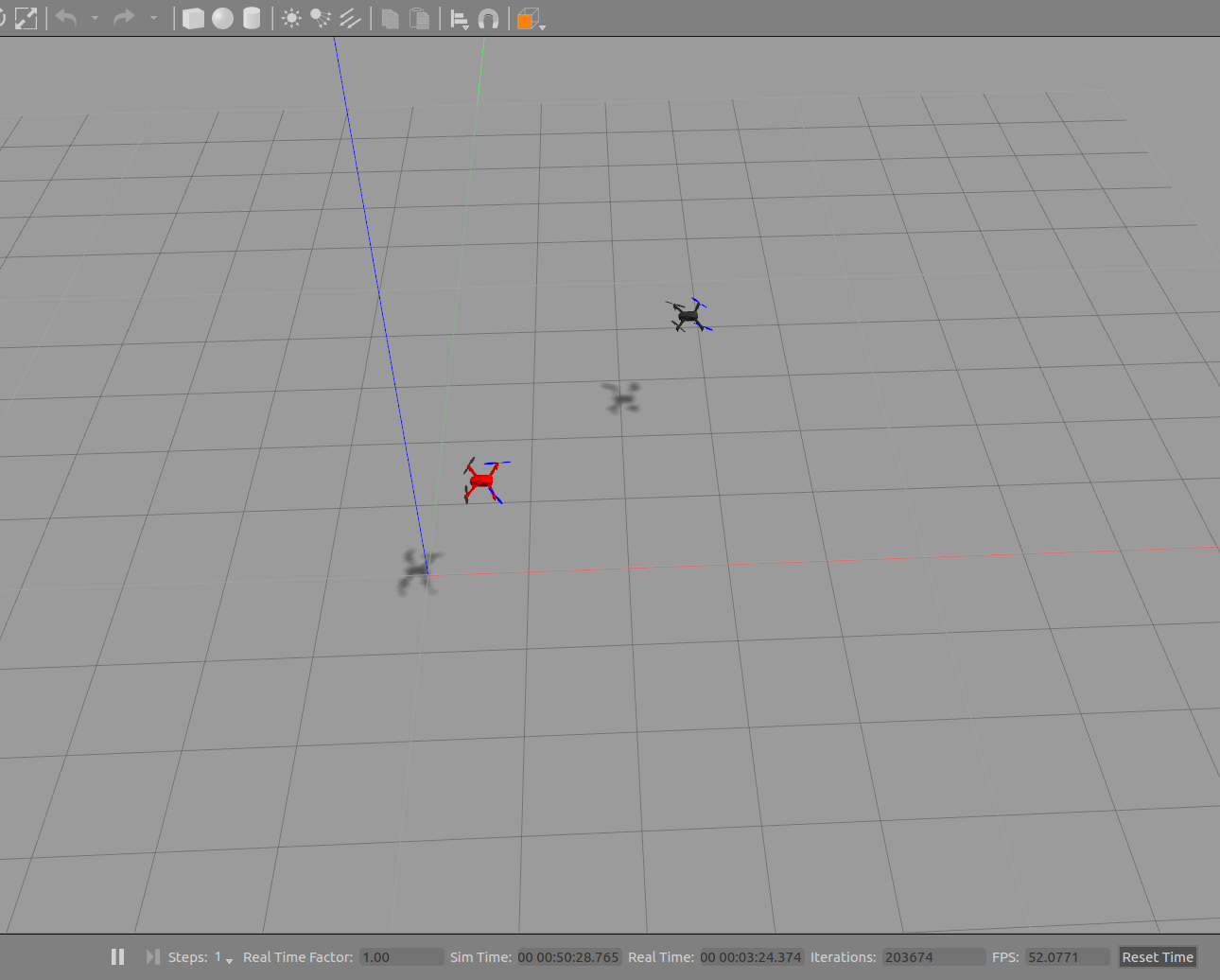}
	\caption{The two-drone setup in Gazebo}
	\label{fig:gazebo_setup}
\end{figure}
We simulated our framework on a two-quadrotor system with the robot operating system (ROS) Gazebo software which provides a realistic simulation environment. We show a snapshot of the simulation setup in Fig.~\ref{fig:gazebo_setup} which consists of an anchor robot $\Rz$ (red drone) and a tag robot $\Ro$ (black drone). We assumed that the robots $\Rz,\Ro$ always move at constant altitudes $z_{0},z_{1}$, respectively. In all simulations, we set the inter-anchor distances $a=0.44\text{m}$ and the desired heights of the robots $z_{0}=z_{1}=2\text{m}.$ We used the Pixhawk controller tools in Gazebo to control the internal dynamics of the quadrotors. We set the Pixhawk controller to the velocity mode which controls the internal dynamics of the quadrotor so as to maintain the quadrotor velocities at given set-points. We considered two scenarios to evaluate the localization performance. In both scenarios, initially the quadrotors hovered and were stabilized at the altitudes $z_{0},z_{1}$. Next, we gave the velocity commands and ran the filtering algorithm concurrently. We used the ``low-variance resampling'' method \cite{ProbabilisticRob_05} in the filtering algorithm.

\subsection{Case~1: Externally Actuated Robots}\label{sec:simcase1}
We gave the velocity set-points to the quadrotors externally and analyzed the localization error for various parameter values. In other words, the estimated state $\mathrm{\hat{r}}$ was not fed to the robot controllers (see Fig.~\ref{fig:block}). We set the system frequency to $f=10$Hz and used the following parameter values:
\begin{align*}
\mathrm{v^{i}_{\min}}&=-2\text{m/s},\quad\mathrm{v^{i}_{\max}}=2\text{m/s},\quad
\sigma_{\mathrm{dist}}=0.05\text{m}.
\end{align*}
This particular choice of $\sigma_{\mathrm{dist}}$ stems from the real-time characteristics of the UWB sensors used in experiments in Section~\ref{sec:experiments}. We set the initial locations of the drones such that $\mathrm{r}(0)=\left[-2,2\right]^{\top}\text{m}$. The yaw angles of the drones remained constant during the entire simulation. The initial condition for the particle estimations were uniformly distributed within the following boundary values:
\begin{align*}
\mathrm{\hat{r}_{x}}(0)&=\textsc{uniform}(-4.2,-0.2)\\
\mathrm{\hat{r}_{y}}(0)&=\textsc{uniform}(-0.2,3.8),
\end{align*}
where $\textsc{uniform}(\cdot)$ denotes the uniform distribution. The robots moved with the following velocities (with respect to the global frame):
\begin{align*}
\mathrm{v^{0G}}&=\left[0,0.2\right]^{\top}\text{m/s},~\mathrm{v^{1G}}=\left[0,0.3\right]^{\top}\text{m/s}.
\end{align*}
\begin{figure}[!b]
	\centering
	\includegraphics[trim = 0cm 0cm 0cm 0cm, clip, width=0.4\textwidth]{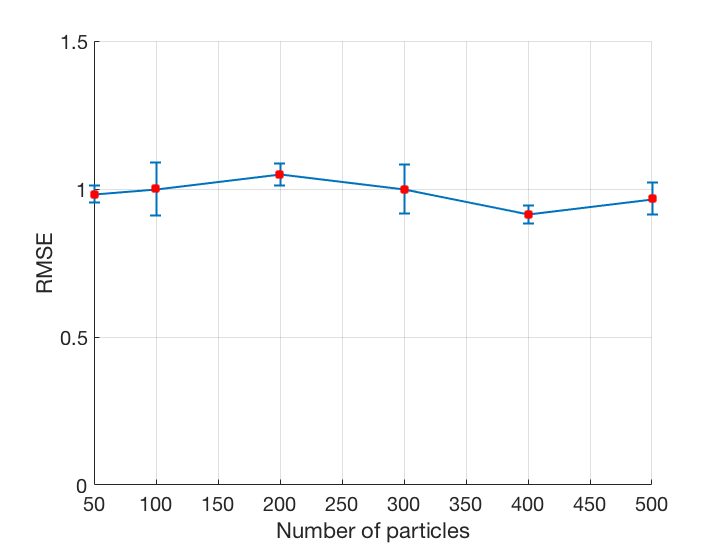}
	\caption{RMSE versus number of particles for $\mathrm{Q_{mot}}=0.5I,~\sigma_{\mathrm{obs}}=0.5$. Each error bar shows the deviation of the five test results for a particular $m$}.
	\label{fig:particle_no}
\end{figure}

\begin{figure}[!t]
	\centering
	\includegraphics[trim = 3cm 1.5cm 3cm 2cm, clip, width=0.45\textwidth]{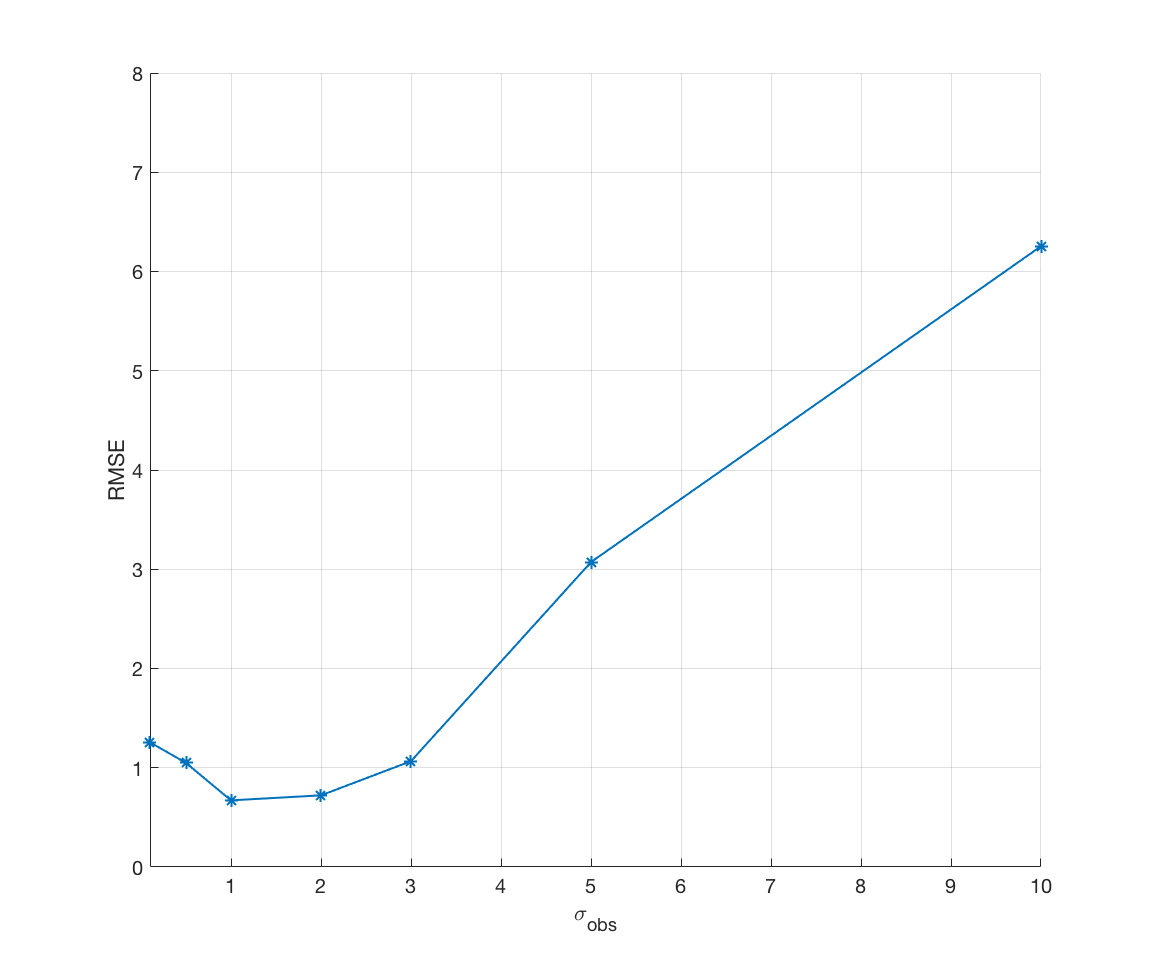}
	\caption{RMSE versus $\sigma_{\mathrm{obs}}$ for $m=200$}
	\label{fig:sigma_obs}
\end{figure}

We tested the performance of the dual MCL algorithm for various numbers of particles (Fig.~\ref{fig:particle_no}) and for various $\sigma_{\mathrm{obs}}$ values (Fig.~\ref{fig:sigma_obs}). We used the root mean square of the relative position error as the performance measure:
\begin{align*}
e_{\mathrm{RMSE}}&=\left(\sum_{i=1}^{m}\frac{1}{m}\|\mathrm{r}-\mathrm{\hat{r}}\|^{2}\right)^{1/2}.
\end{align*}
We performed five simulation runs for each choice of $m$. We show the average (red square) and standard deviation (blue vertical bars) of the errors in Fig.~\ref{fig:particle_no}. We observed that the performance of the algorithm does not vary significantly with the increasing number of particles. Remarkably, we expect this result for our particular dual MCL algorithm because the parameter that greatly affects the particle distribution is not the number but the variance of the  particles around the measurement $\mathrm{r}_{k}^\mathrm{meas}$. Since even a small number of particles cover a sufficient region around the measurement $\mathrm{r}_{k}^\mathrm{meas}$, we do not observe a significant degradation in performance as the number of particles reduces.

If the deviation of the particle distribution is too high, more particles would likely be assigned small weights. On the other hand, if the deviation is too low, the majority of the particles would be condensed around the measurement $\mathrm{r}_{k}^\mathrm{meas}$, which would lead to a performance degradation. We tested seven $\sigma_{\mathrm{obs}}$ values for a particular set of parameters, setting $m=200$ (Fig.~\ref{fig:sigma_obs}). The setting $\sigma_{\mathrm{obs}}\approxeq1$m yielded the best performance.

\begin{figure}[!t]
	\centering
	\includegraphics[trim = 3cm 1cm 3.0cm 3cm, clip, width=0.48\textwidth]{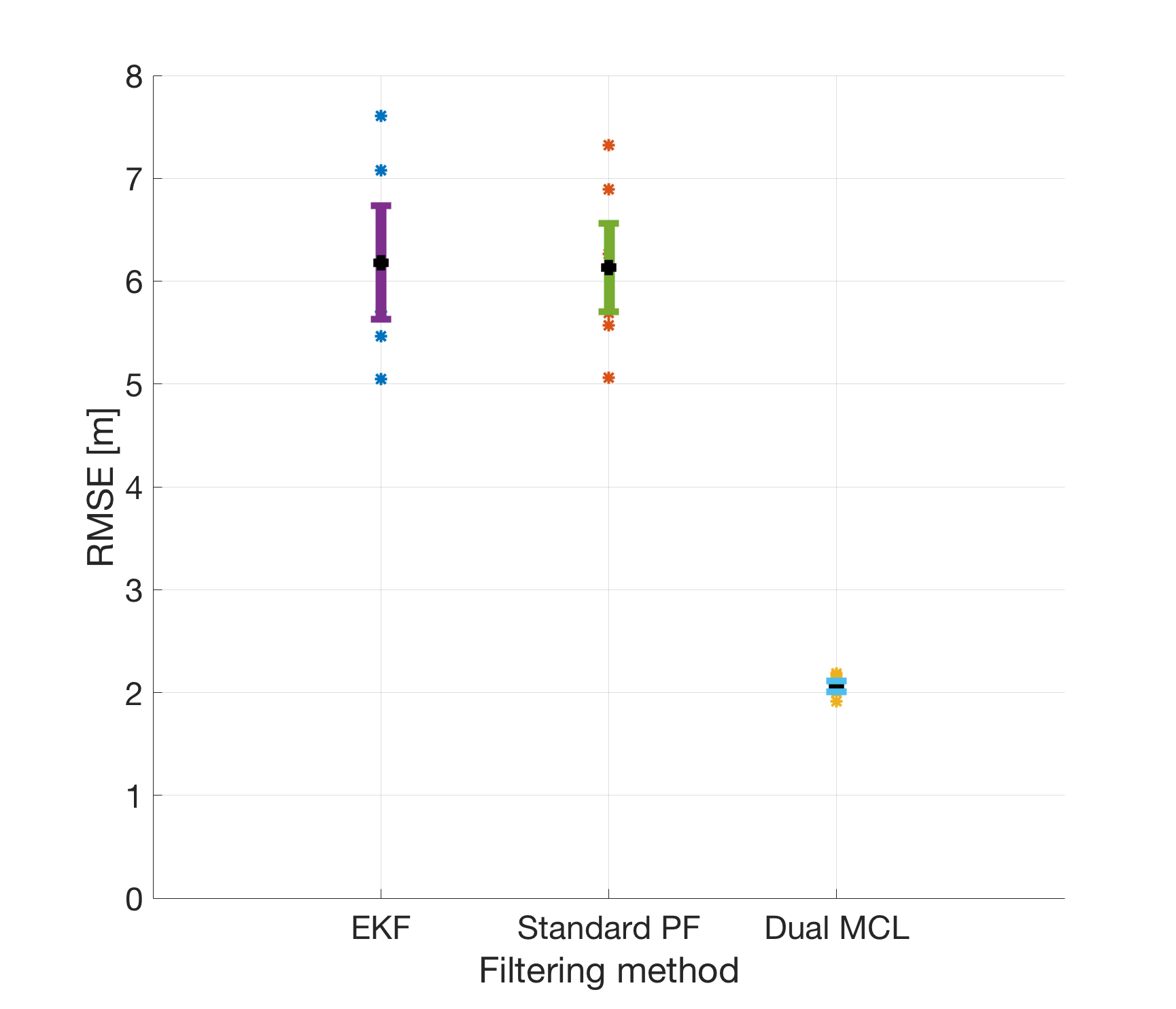}
	\caption{RMSE for the standard PF, EKF, and the proposed dual MCL algorithms. The data points for the standard PF and the proposed algorithms represent the RMSE for different $\sigma_{\mathrm{obs}}$ values. The data points for the EKF algorithm represents the RMSE for different motion and observation covariance matrices. The vertical bars and their centers represent the deviations and average values of RMSEs, respectively.}
	\label{fig:rmse_all}
\end{figure}

Furthermore, we compared the performances of the proposed dual MCL algorithm with the standard PF and EKF algorithms for the case of an agile robot $\Ro$ (Fig.~\ref{fig:rmse_all}). We set $\mathrm{v^{0G}}=\left[0,0.2\right]$m/s, $\mathrm{v^{1G}}=\left[4\kappa,0.3\right]$m/s, where $\kappa=\{-1,1\}$ is a switching function based on time. We simulated the proposed algorithm and the standard PF algorithm for various $\sigma_{\mathrm{mot}}$ values. We tested the EKF algorithm of \cite{GulerCCTA} for five sets of parameter values. We used relatively high covariance matrices in the EKF algorithm because robot $\Ro$ was able to show agile motion behavior with a relatively high maximum speed. Although our algorithm yielded $2$m/s RMSE on average, it outperformed the standard PF and EKF algorithms which could not capture the agile motion of robot $\Ro$ and estimate the relative position within acceptable bounds. Also, we observed that the standard PF and EKF estimations tracked the actual relative positions with time delay. 

\subsection{Case~2: Localization-based Formation Control}\label{sec:simcase2}
We tested the localization performance in a feedback control system on robot $\Rz$. We sent the velocity set-points to robot $\Ro$ while robot $\Rz$ was to maintain the relative position $\mathrm{r}$ at a desired constant value by utilizing the estimate $\mathrm{\hat{r}}$. Similar to Case~1, the quadrotors first hovered and were stabilized at the pre-defined altitudes $z_{0},z_{1}$. Then, both quadrotors started to move simultaneously. To imitate the characteristics of the DecaWave UWB sensors, we set the loop rate $f=3.3$Hz and the distance noise variance $\sigma_{\mathrm{dist}}\in\{0.05,0.1\}$m. We used a simple proportional controller for formation maintenance as follows:
\begin{align}\label{eqn:proportionalCtr}
\mathrm{v_{0x}^{des}}&=K_{v}e_{x},\quad \mathrm{v_{0y}^{des}}=K_{v}e_{y},
\end{align}
where $e_{x}=\mathrm{r_{x}^{des}}-\mathrm{\hat{r}_{x}},~e_{y}=\mathrm{r_{y}^{des}}-\mathrm{\hat{r}_{y}}$. We found the $K_{v}$ value which gives the best observed performance empirically. To avoid chattering around $e_{x}=e_{y}=0$, we set $\mathrm{v_{0x}^{des}}=\mathrm{v_{0y}^{des}}=0$ in the region $\mathrm{\hat{r}_{i}}\in[-0.2,0.2]$ in both axes.

\begin{figure}[!t]
	\centering
	\includegraphics[trim = 3.5cm 0cm 4cm 2cm, clip, width=0.48\textwidth]{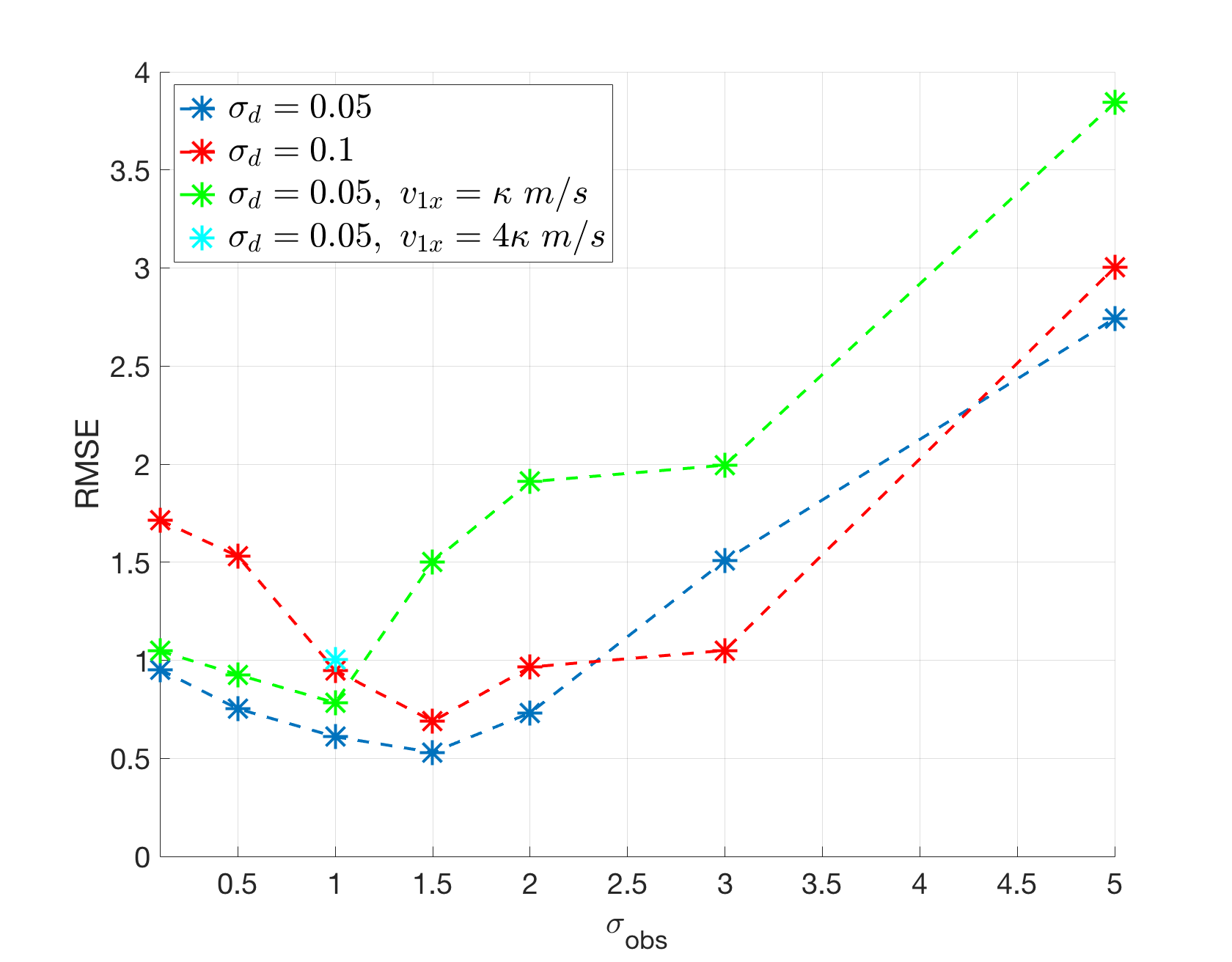}
	\caption{RMSE versus $\sigma_{\mathrm{obs}}$ for $N=400$ and for different distance noise levels}
	\label{fig:dual_LF_sobs_all}
\end{figure}

We set $m=400$ and tested the performance of our algorithm in the formation control application for seven $\sigma_{\mathrm{obs}}$ values (Fig.~\ref{fig:dual_LF_sobs_all}). We analyzed effects of the distance noise level and robot $\Ro$'s velocity profiles on the performance. We observed similar characteristics in RMSE for different $\sigma_{\mathrm{obs}}$ values: The errors diminished around the region $\sigma_{\mathrm{obs}}\in(1,1.5)$ and increased as $\sigma_{\mathrm{obs}}$ increased or approached zero. The blue and red lines depict the performances for noise levels $\sigma_{\mathrm{dist}}=0.05$ and $\sigma_{\mathrm{dist}}=0.1$, respectively. The errors with $\sigma_{\mathrm{dist}}=0.1$ were larger than the errors with $\sigma_{\mathrm{dist}}=0.05$ for $\sigma_{\mathrm{obs}}<2$ as expected, because the particles dispersed around a small neighborhood of the measurement $\mathrm{r^{meas}}_k$ for $\sigma_{\mathrm{obs}}<2$. We observe the minimum errors for all cases around $\sigma_{\mathrm{obs}}=1.5$.

The green line depicts the error when $\mathrm{v_{1x}}=\kappa$, where $\kappa=\{-1,1\}$ is a switching function based on time. This velocity profile demonstrates an agile motion behavior. We observed that $e_{\mathrm{RMSE}}$ increased compared to the straight $\mathrm{v_{1}}$ case. Yet, the error was minimum for $\sigma_{\mathrm{obs}}\cong1$ similar to the previous velocity profiles. We also tested our algorithm's performance for a more agile motion behavior by setting $\mathrm{v_{1x}}=4\kappa$. Our algorithm performed well and captured the periodic motion of the tag robot (light blue star in Fig.~\ref{fig:dual_LF_sobs_all}).

Remarkably, the RMSE errors show the same characteristics with Case~1 result (Fig.~\ref{fig:sigma_obs}) in that the errors diminish as $\sigma_{\mathrm{obs}}$ approaches the region $\sigma_{\mathrm{obs}}\in(1,1.5)$.

\setcounter{equation}{0}
\section{Experiments}\label{sec:experiments}
\subsection{Experimental Setup}\label{sec:expSetup}
\begin{figure}[!t]
	\centering
	\includegraphics[trim = 0cm 0cm 0cm 0cm, clip, width=0.45\textwidth]{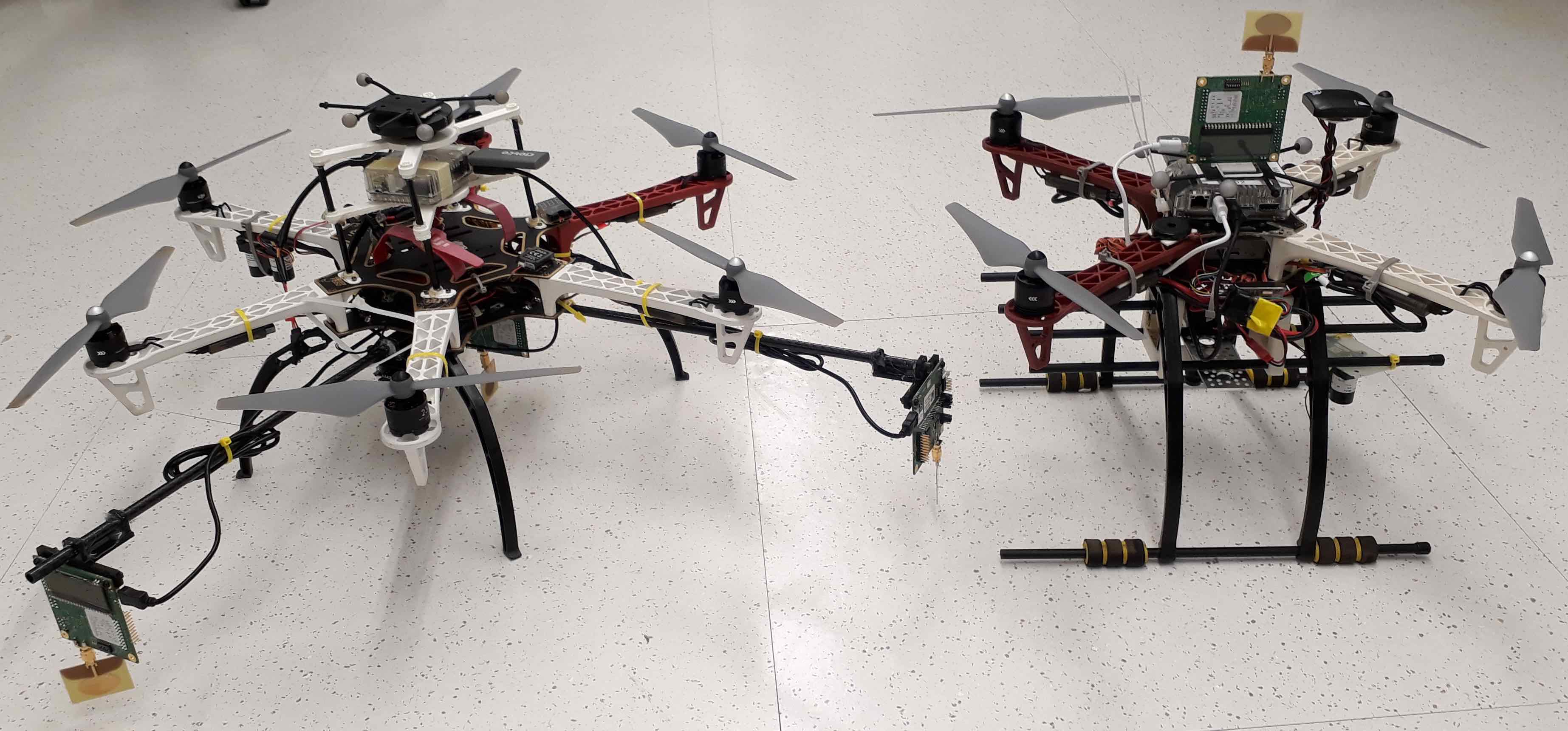}
	\caption{The hexacopter and the quadrotor used in the experiments}
	\label{fig:drones}
\end{figure}

We performed experiments with two drones, a hexacopter equipped with three UWB anchors and a quadrotor equipped with a single UWB sensor (Fig.~\ref{fig:drones}). The hexacopter was to estimate the relative position to the quadrotor. Each drone was equipped with a laser range finder for precision altitude control, camera-based flow sensor for hovering and velocity estimation, a flight controller for drone low-level control, and a low-power Linux computer for localization and filtering computations (Table~\ref{table:drone_setup}). Particularly, each drone used a Pixhawk flight controller\footnote{\url{https://pixhawk.org}} running a PX4 open-source autopilot firmware to provide attitude stability and velocity tracking. The flight controller used the PX4Flow sensor \cite{px4flow} to provide accurate velocity feedback and hovering. We used an onboard Odroid XU4 computer\footnote{\url{https://www.hardkernel.com/main/products}} as a high-level controller to send the velocity set-points to the flight controller and to execute the localization and filtering algorithms.
\begin{table}[t!]
	\centering
	\caption{Drone Components}
	\label{table:drone_setup}
	\begin{tabular}{|c|p{0.6\linewidth}|}
		\hline
		\textbf{Component} & \textbf{Description}                                                                                               \\ \hline
		Airframe           & Hexacopter of diameter 50 cm. Custom 3D printed quadrotor frame of diameter 35 cm                                                                                       \\ \hline
		Flight Controller  & Pixhawk with PX4 autopilot firmware                                                                             \\ \hline
		Range Finder       & LiDAR Lite v3 for precision altitude measurements                                                                  \\ \hline
		PX4FLOW            & A camera flow sensor for hover stabilization and velocity feedback                                                 \\ \hline
		Onboard Computer   & Odroid XU4 installed with Ubuntu~16 and ROS Kinetic for high-level computations \\ \hline
		UWB sensors        & Decawave TREK1000 for localization                                                                                                   \\ \hline
	\end{tabular}
\end{table}

We conducted several indoor and outdoor experiments. The videos of some experiments are available online\footnote{\url{https://drive.google.com/drive/folders/1-IjM1TqZSzGbNfJdEvGRd5kBT9_zjJQb?usp=sharing}}. Since we assumed that the drones maintained constant heading during the entire operation, we initiated the drones at the desired configuration and set the attitude controller so as to maintain the yaw angles of the drones constant during the operation. Notably, this approach does not restrict the motion capabilities of the anchor drone, i.e., a drone can reach the entire plane with a constant heading. The axes of the desired configuration formed the virtual $x$-$y$ axes of the localization algorithm as described in section~\ref{sec:proposedAlgo}. This frame was set as the global frame $\Fcal_{G}$ as well. The test procedure consisted of two stages. In the first stage, we brought the drones to certain locations and altitudes manually. Then, we switched to the autonomous mode and ran the localization algorithm. In the autonomous mode, the quadrotor (robot $\Ro$) moved based on pre-defined velocity set-points, and the hexacopter (robot $\Rz$) moved based on either external inputs or the localization algorithm feedback. To avoid occlusions between the UWB anchors and the sensor, the drones flew at different altitudes. The altitude controllers were set to maintain the altitude of the drones at desired values.

\subsection{UWB Sensor Calibration} \label{sec:UwbCharacterization}
We followed a common procedure to calibrate the DecaWave UWB sensors as explained in \cite{GulerCCTA}. First, we recorded a large amount of distance data for the LOS case for each anchor. Then, we found the bias and noise variance for the anchors.

\subsection{Indoor Experiments}\label{sec:indoor}
We performed three test procedures and recorded the ground truth data from a motion capture system. For high precision, the control algorithms maintained the yaws and altitudes of the drones constant by the aid of the motion capture data. 
\begin{figure}[!b]
	\centering
	\includegraphics[trim = 5cm 3.2cm 5cm 4cm, clip, width=0.4\textwidth]{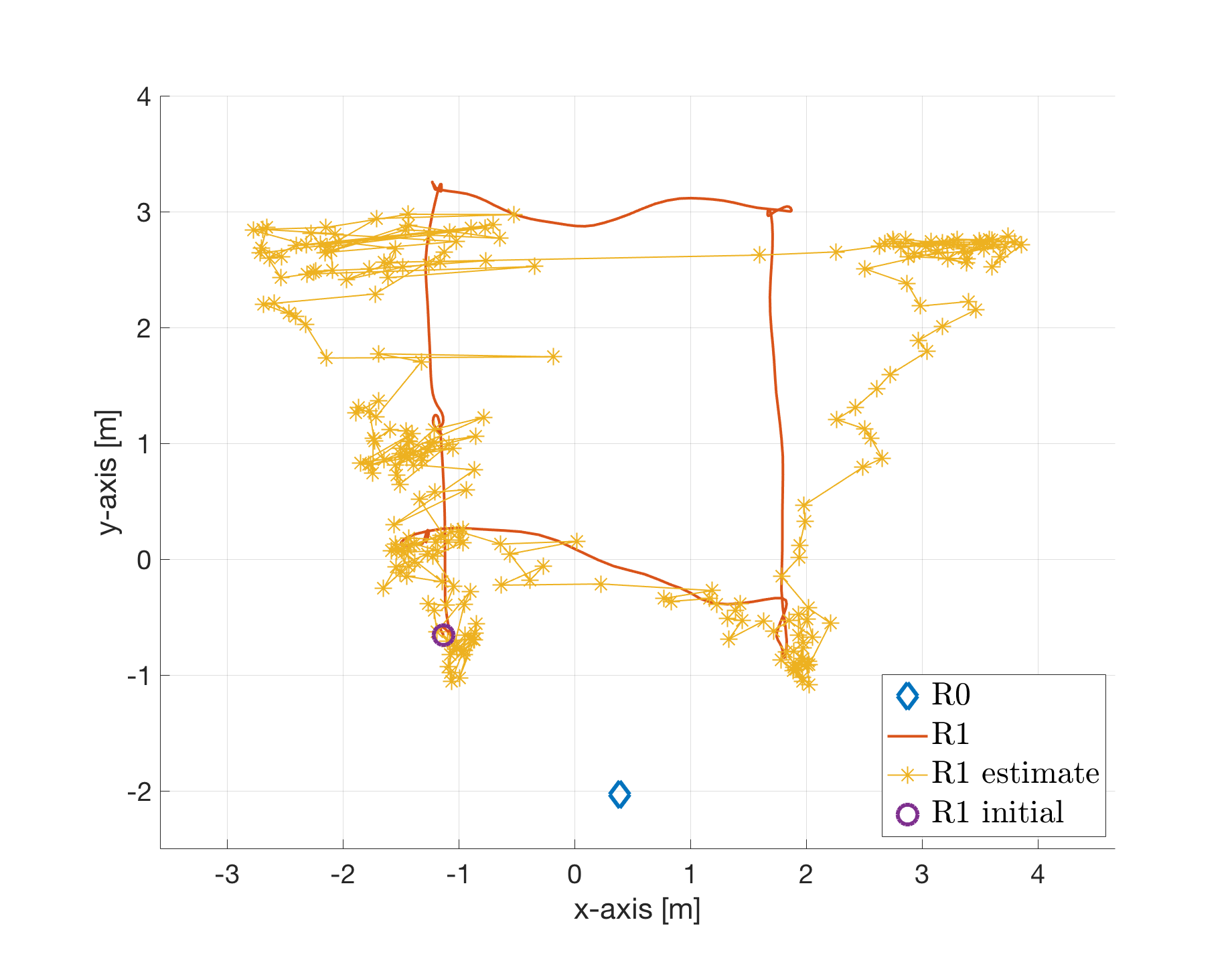}
	\caption{The quadrotor moved on a square shape trajectory whereas the hexacopter remained stationary. The yellow line represents the state estimate.}
	\label{fig:indoor_square_plane}
\end{figure}
\begin{figure}[!t]
	\centering
	\includegraphics[trim = 2cm 0cm 1.7cm 1cm, clip, width=0.48\textwidth]{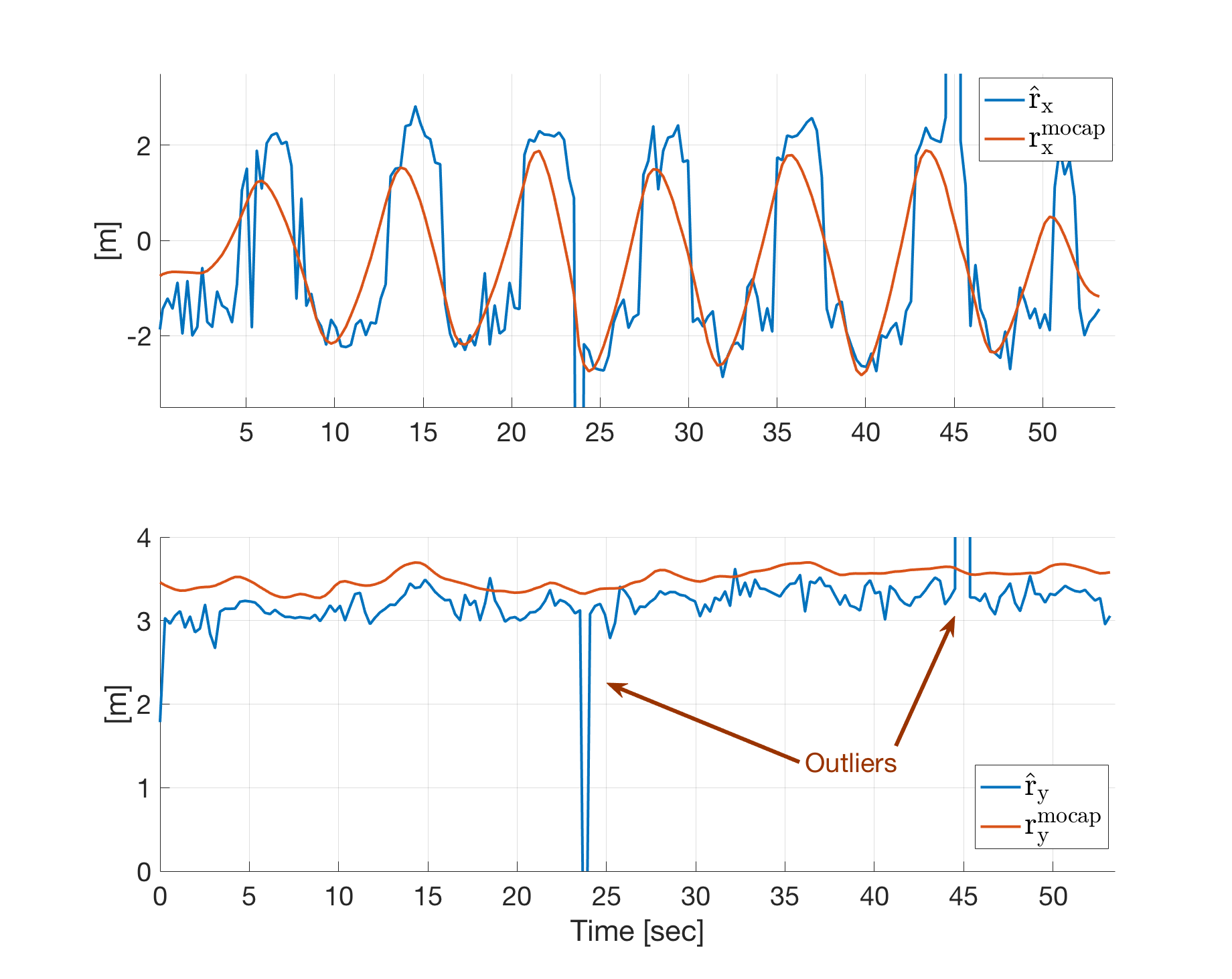}
	\caption{Relative position estimates and ground truth. The quadrotor moved with a periodic velocity profile ($\pm0.4$m/s) whereas the hexacopter remained stationary.}
	\label{fig:indoor_pure_0p4}
\end{figure}

\subsubsection{Procedure~1}\label{sec:IndoorProc1}
We performed two tests by keeping the hexacopter stationary. In the first test, the quadrotor traversed an almost square shape with piecewise constant speeds (Fig.~\ref{fig:indoor_square_plane}). Although the initial state estimate $\mathrm{\hat{r}}(0)$ was away from its actual value, our algorithm yielded a reasonable accuracy. In the second test, the quadrotor moved with speed $\mathrm{v^{1G}}=\left[0.4\kappa,0\right]^{\top}\text{m/s}$, where $\kappa=\{-1,1\}$ switched with a period of $6$sec, thereby moving the drone in a periodic fashion in the $x$-axis (Fig.~\ref{fig:indoor_pure_0p4}). Although there is a small offset between the actual and estimated values, the algorithm caught the periodic profile of the state in the $x$-axis.
\begin{figure}[!t]
	\centering
	\includegraphics[trim = 3cm 3cm 3cm 3cm, clip, width=0.45\textwidth]{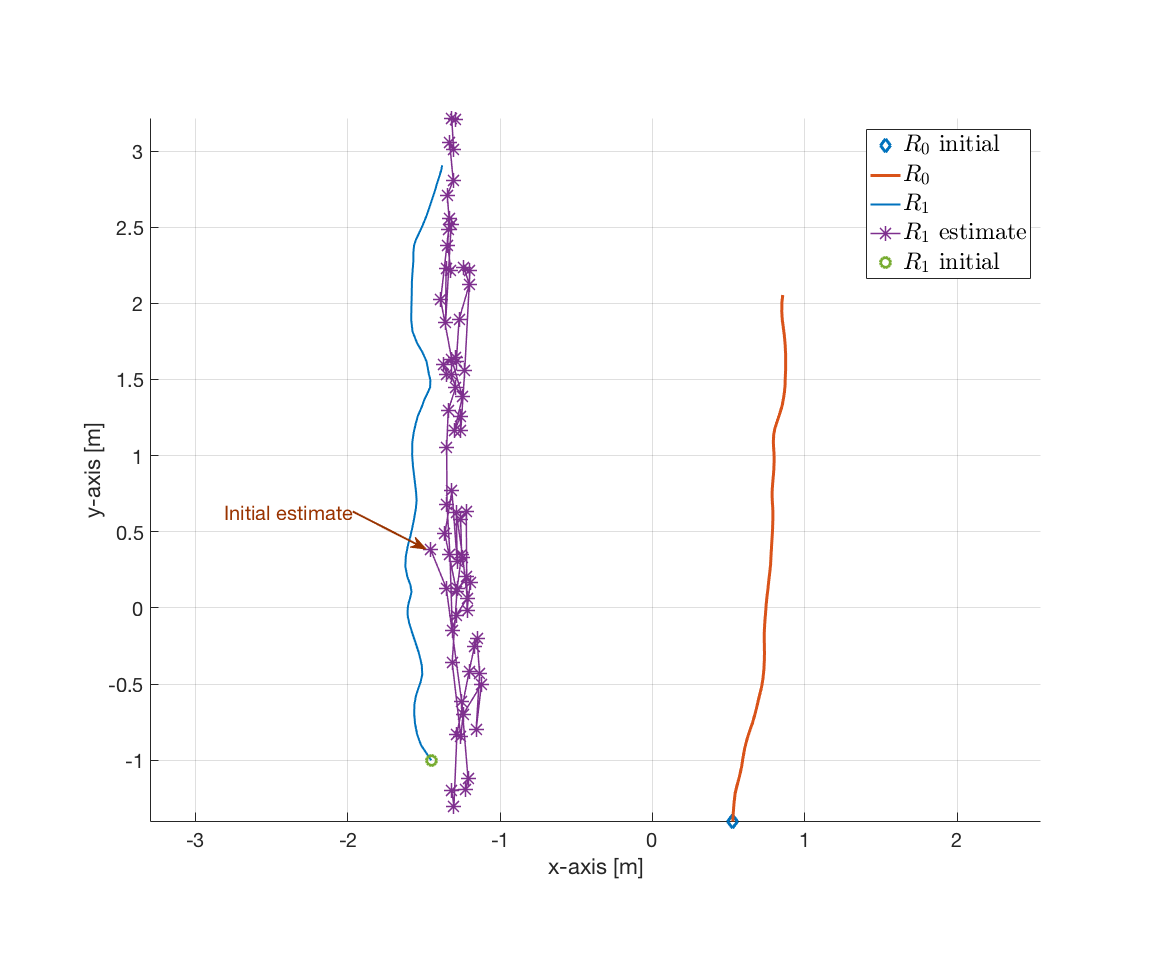}
	\caption{The estimated location of robot $\Ro$ (purple) and the trajectory of both robots}
	\label{fig:indoor_bothmvng_1_plane}
\end{figure}

\subsubsection{Procedure~2}\label{sec:IndoorProc2}
In this set of tests, both robots moved with external control inputs. In this case, we expect a degradation in the estimation performance because the inaccuracies in not only motion of robot $\Ro$ but also motion of robot $\Rz$ adversely affect the performance. Also, the distance reading accuracy of the UWB sensors are affected by motions. Although there was a big difference in the initial values of the actual and estimated relative positions, we observed a sufficient level of precision in this test (Fig.~\ref{fig:indoor_bothmvng_1_plane}).
\begin{figure}[!t]
	\centering
	\includegraphics[trim = 3cm 2cm 3cm 2cm, clip, width=0.45\textwidth]{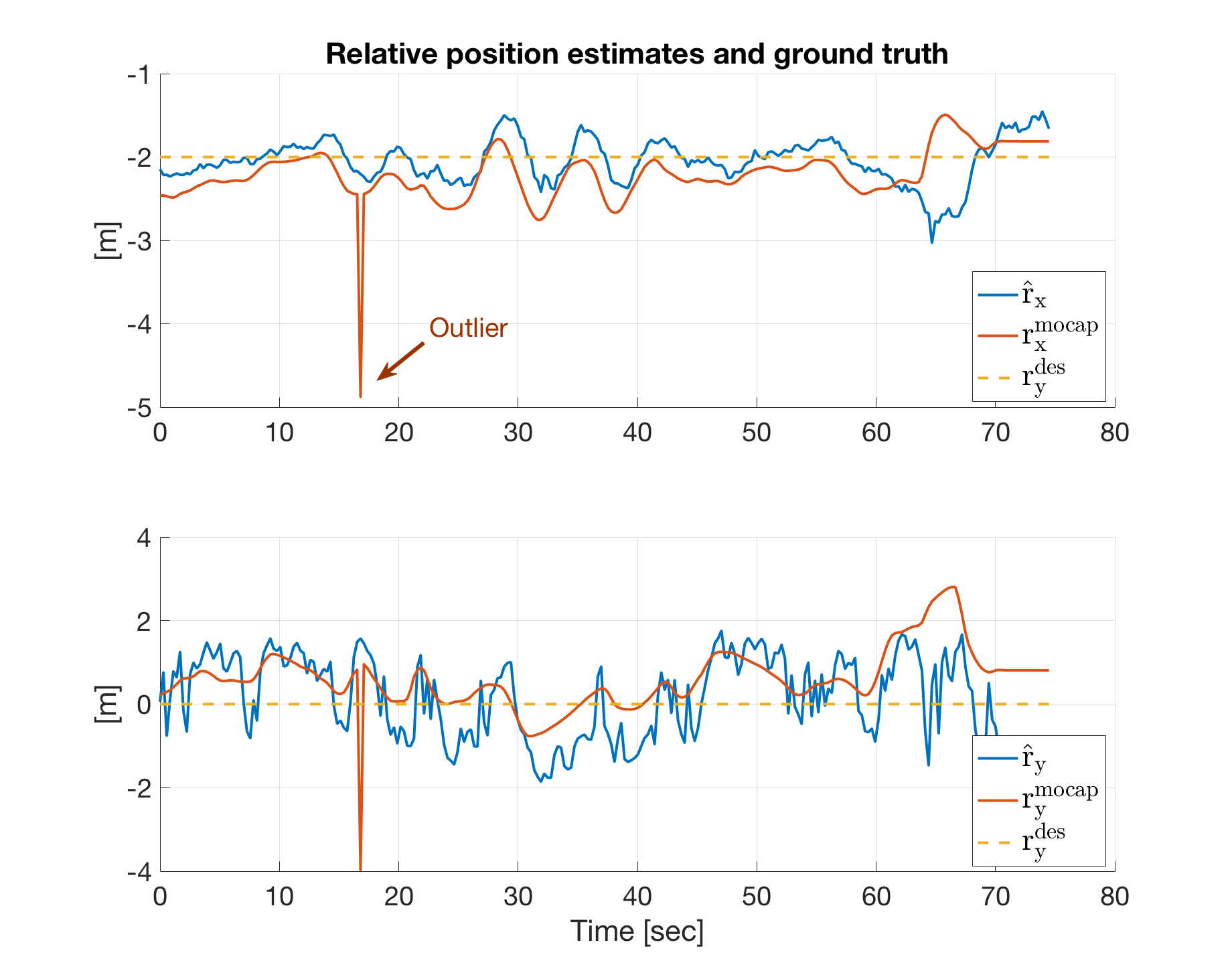}
	\caption{Formation control based on localization feedback experiment. The blue and red lines show the relative pose estimates and the ground truth. The yellow lines are the desired locations in both axes.}
	\label{fig:indoor_lf_periodic_2}
\end{figure}

\subsubsection{Procedure~3}\label{sec:IndoorProc3}
Similar to the algorithm presented in Section~\ref{sec:simcase2}, here we implemented a simple formation control algorithm on robot $\Rz$ (Fig.~\ref{fig:indoor_lf_periodic_2}). We used the proportional controller \ref{eqn:proportionalCtr} with $K_{v}=1$. We demonstrate the estimated relative position, the ground truth data, and the desired relative position in Fig.~\ref{fig:indoor_lf_periodic_2}.
\begin{table}[t!]
	\caption{RMSE Errors of Indoor Experiments}
	\label{table:rmseIndoor}
	\begin{center}
		\begin{tabular}{| p{0.4\linewidth} | c | c | c |}
			\hline
			\textbf{Experiment} & $\mathbf{\sigma_{\mathrm{obs}}}$ (m) & $\mathrm{v^{i}_{\max}}$ (m/s) & $e_{\mathrm{rmse}}$ (m) \\ \hline
			Externally actuated 
			
			($\Rz$: stationary, $\Ro$: square) & $0.2$ & $0.1$ & $1.1635$ \\ \hline
			Externally actuated
			
			($\Rz$: stationary, $\Ro$: periodic) & $0.4$ & $1$ & $1.4225$ \\ \hline
			Externally actuated
			
			($\Rz$: straight, $\Ro$: straight) & $0.2$ & $1$ & $0.70365$ \\ \hline
			Formation control
			
			($\Rz$: controlled, $\Ro$: straight) & $0.8$ & $0.5$ & $1.1716$ \\ \hline
			Formation control
			
			($\Rz$: controlled, $\Ro$: periodic) & $0.5$ & $0.5$ & $1.356$ \\
			\hline
		\end{tabular}
	\end{center}
\end{table}

We show the RMSE of all indoor experiments in Table~\ref{table:rmseIndoor}. We observed errors at similar levels to the simulations.

\subsection{Outdoor Experiments}\label{sec:outdoor}
We conducted outdoor experiments for both the externally actuated case and the formation control case. We used an optical flow sensor on each drone for the planar motion control. We set the desired altitudes $z_{0}=2$m, $z_{1}=1$m and used a laser range sensor to stabilize the altitudes of the drones. We acquired the ground truth data from GPS sensors on-board. We followed the same test procedure with the indoor case: We first stabilized the drones at the desired altitudes, then ran the localization algorithm and sent the velocity set-points.

\subsubsection{Procedure~1}\label{sec:OutdoorProc1}
\begin{figure}[!t]
	\centering
	\includegraphics[trim = 2.7cm 1cm 3cm 2cm, clip, width=0.4\textwidth]{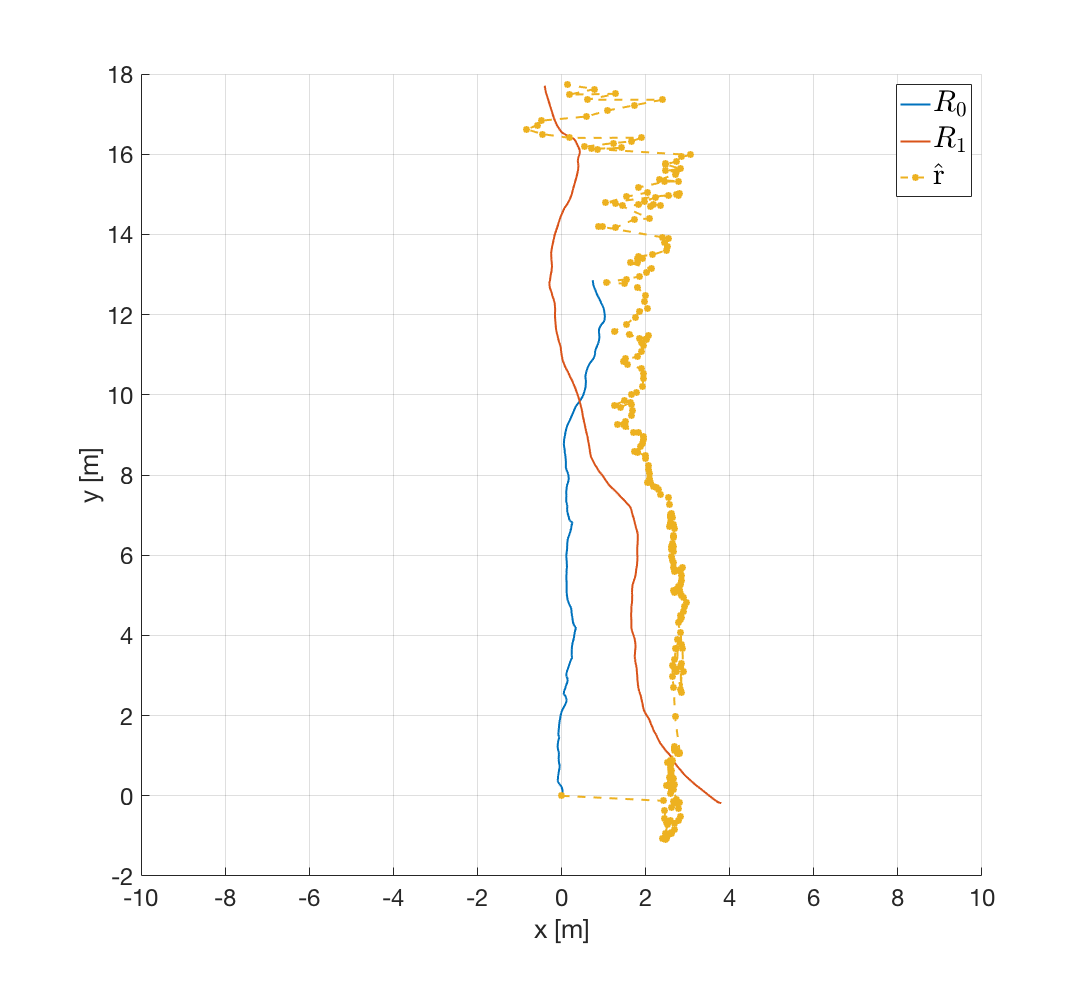}
	\caption{Outdoor experiment: The drones' trajectories and the relative position estimate; $\mathrm{v^{0G}}=\left[0,0.2\right]^{\top}$m/s,
		$\mathrm{v^{1G}}=\left[0,0.3\right]^{\top}$m/s.}
	\label{fig:ex1_plane}
\end{figure}

In this set of tests, we moved the hexacopter with velocity $\mathrm{v^{0G}}=\left[0,0.2\right]^{\top}\text{m/s}$. In the first test, we set $\mathrm{v^{1G}}=\left[0,0.3\right]^{\top}\text{m/s}$. We show the drones' trajectories and the relative position estimates in Fig.~\ref{fig:ex1_plane}. The estimate was biased from the ground truth data. In the second test, we set $\mathrm{v^{1G}}=\left[0,1.2\kappa\right]^{\top}\text{m/s}$, where $\kappa=\{-1,1\}$ switched with a period of $6$sec.
\begin{figure}[!t]
	\centering
	\includegraphics[trim = 2.7cm 1cm 3cm 2cm, clip, width=0.4\textwidth]{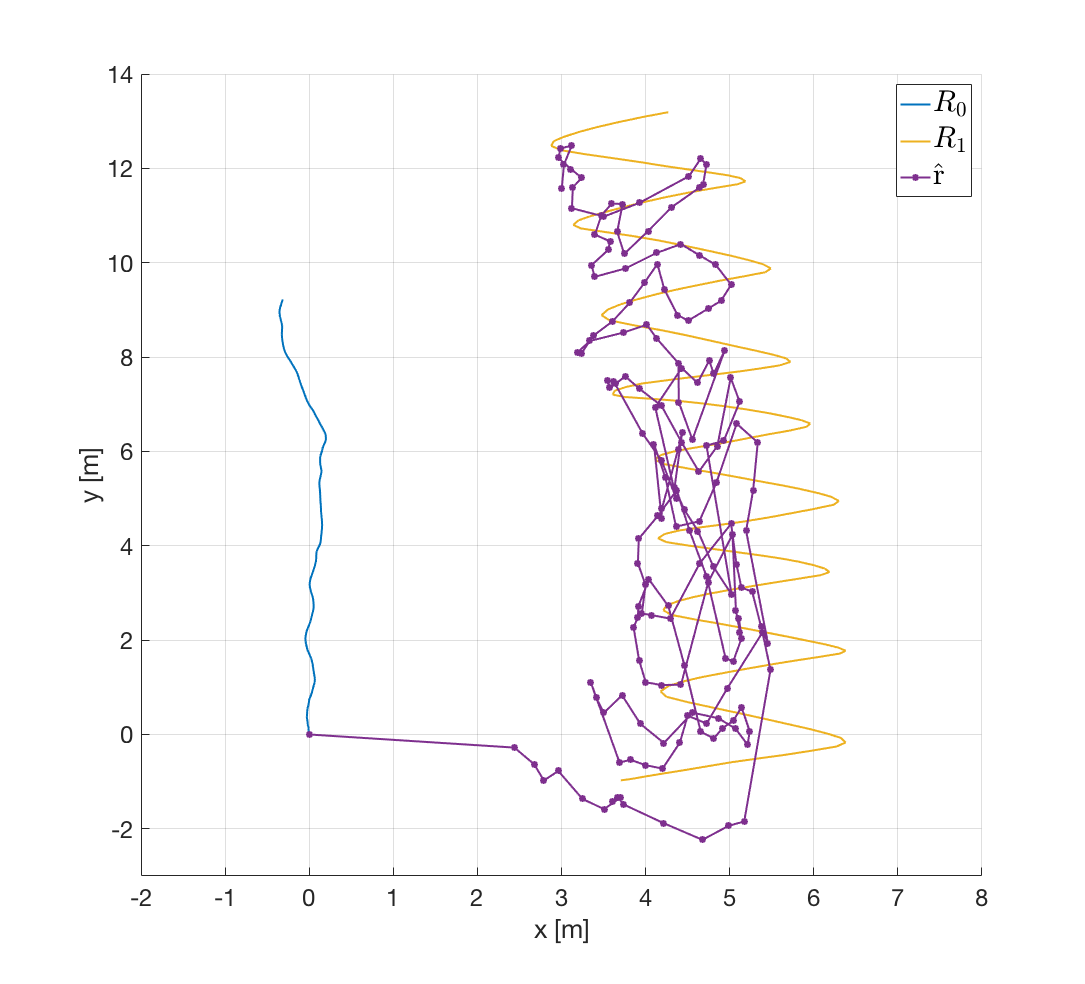}
	\caption{Outdoor experiment: The drones' trajectories and the relative position estimate; $\mathrm{v^{0G}}=\left[0,0.2\right]^{\top}$m/s,
		$\mathrm{v^{1G}}=\left[0,1.2\kappa\right]^{\top}$m/s.}
	\label{fig:ex3_plane}
\end{figure}

\subsubsection{Procedure~2}\label{sec:OutdoorProc2}
\begin{figure}[!t]
	\centering
	\includegraphics[trim = 3cm 1.5cm 3cm 2.5cm, clip, width=0.4\textwidth]{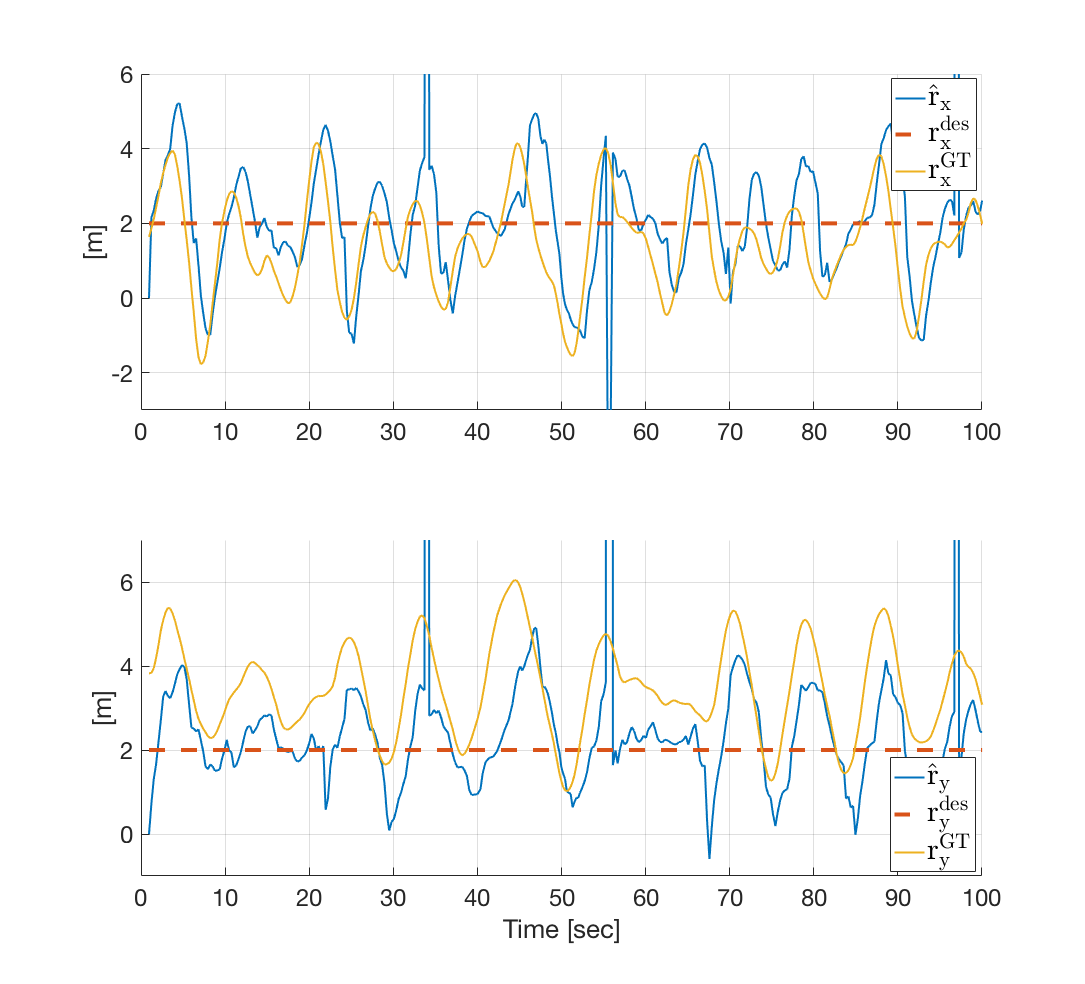}
	\caption{Outdoor experiment: Formation control based on localization feedback.}
	\label{fig:ex4_est}
\end{figure}

We performed formation control experiments. We moved robot $\Ro$ with velocity $\mathrm{v^{1G}}=\left[0,1.2\kappa\right]^{\top}\text{m/s}$, where $\kappa=\{-1,1\}$ switched with a period of $6$sec. The hexacopter (robot $\Rz$) was to maintain the relative position at the desired value $\mathrm{r^{des}}=\left[2,2\right]^{\top}$m with the proportional controller \eqref{eqn:proportionalCtr}. We show the relative position estimate in Fig.~\ref{fig:ex4_est}. Here, we emphasize that the oscillation shown in both axes mainly stem from the controller mechanism which was chosen a simple proportional controller. Importantly, our algorithm captured the agile maneuvers of the tag drone within an acceptable bound.

\section{Discussion on Results}\label{sec:discussion}
We presented simulation and indoor and outdoor experiment results for two common cases, namely, localization in externally actuated robots and localization-based formation control, to demonstrate the effectiveness of our algorithm. Our algorithm yielded sufficient accuracy in most simulation runs. We used less than $500$ particles in all simulations. Furthermore, we showed that our algorithm can yield sufficient accuracy even in tracking agile motion behaviors by utilizing only $400$ particles. We now note some important practical aspects.

Firstly, as a natural outcome of the particular problem setup, we would like to localize a moving robot on another moving robot. Furthermore, the localized robot is allowed to show aggressive behavior. This problem is harder to tackle compared to the cases where a moving robot localizes itself by taking measurements from stationary landmarks. We emphasize that not only the distance measurement noises, but also the motion inaccuracies such as actuator malfunctions affect the performance of our algorithm. We argue that the high level of errors in some simulations arise mainly from this issue.

Secondly, the loop rate, or the system frequency, has a major impact on the performance. It is common to use high-frequency sensors in drone applications, such as IMU with $1000$Hz data rate. However, our UWB sensors generate data at around $3.5$Hz, and we aimed to imitate the real-life scenario in our simulations. Therefore, we set the loop rate at $3.33$Hz in Case~2, which corresponds to a $3$s interval between two successive time steps. While this setting gives a sufficient amount of time for filtering, the drone's control mechanism works much better in higher frequencies. We argue that the error levels in Fig.~\ref{fig:dual_LF_sobs_all} can be reduced dramatically by increasing the sensor frequency if possible, and hence, the loop rate.

Thirdly, we emphasize that we combine solutions for two separate problems in Case~2, localization and motion control, by feeding the estimated state vector to the motion control algorithm. Admittedly, the solution to the motion control objective affects the performance of the overall algorithm. Therefore, the particular setting in Case~2 is a combination of the two objectives. Improvement of the motion control performance based on localization feedback is a separate task and beyond the scope of this work. We argue that a different formation control algorithm may yield a better tracking performance.

Fourthly, unlike the standard particle filter and EKF, our algorithm does not require an initial guess for the relative position estimate. Therefore, the robots may start or end the localization at any time in an operation. This feature provides great flexibility for some applications such as the kidnapped robot problem \cite{ProbabilisticRob_05}. On the other hand, our algorithm suffers from high noise in distance measurements which causes chattering in the estimation outcome. This chattering could be overcome by an additional smoothing filter at the expanse of losing the capability of  tracking agile robots.

Finally, we believe that the offsets between the actual and estimated tag robot locations in Fig.~\ref{fig:ex1_plane} and \ref{fig:ex3_plane} and the offset between the actual and estimated relative position in $y-$axis in Fig.~\ref{fig:ex4_est} emerge from miscalculations of the ground truth data. Particularly, the estimates in Fig.~\ref{fig:ex1_plane} and \ref{fig:ex4_est} show the same characteristics as the actual data. We plan to employ a differential GPS to acquire more reliable ground truth data in future experiments.

\section{Conclusion}\label{sec:conclusion}
Motivated by the need for a reliable and versatile multi-robot localization solution, we have designed an on-board UWB localization framework for a two-robot system. Our framework utilizes the UWB distance measurements and motion models of the robots to generate an estimate of the inter-robot relative position in real-time. We exploit the non-linear structure of the dual MCL algorithm to generate accurate estimates for a broad class of tag robot velocity profiles including agile maneuvers. Remarkably, our framework runs on board the anchor robot in real time without any need for a central computational unit such as a ground station. Also, our framework does not employ an explicit communication structure. Therefore, our framework provides a flexible multi-robot localization solution for both indoor and outdoor operations. Extensive simulation and experimental studies proved the reliability and repeatability of our framework. To the best of our knowledge, this work is the first to represent a real-time, on-board multi-robot localization framework tested on a two-drone setup both in indoor and outdoor experiments.

In future, we plan to extend our framework to three-dimensional scenarios by adding an extra UWB sensor to the anchor drone. Furthermore, we plan to study various advanced control techniques to improve the formation control performance. Particularly, we believe that model predictive control can improve the tracking accuracy for the particular estimation profiles we obtained.

\bibliographystyle{IEEEtran}
\bibliography{IEEEabrv,UWB_Journal}

\end{document}